\documentclass[acmtog,authorversion]{acmart}
\acmSubmissionID{337}

\usepackage{booktabs} 

\setcitestyle{square}

\usepackage[ruled]{algorithm2e} 

\SetAlFnt{\small}
\SetAlCapFnt{\small}
\SetAlCapNameFnt{\small}
\SetAlCapHSkip{0pt}
\IncMargin{-\parindent}

\acmJournal{TOG}
\acmVolume{38}
\acmNumber{4}
\acmArticle{75}
\acmYear{2019}
\acmMonth{7}

\setcopyright{acmcopyright}

\acmDOI{10.1145/3306346.3322999}

\newcommand{\etal}{et al.}
\usepackage{color}
\usepackage{soul}
\usepackage{breqn}
\usepackage{booktabs}
\usepackage{rotating}
\usepackage{xcolor}
\usepackage[symbol]{footmisc}

\definecolor{purple}{rgb}{0.65,0,0.65}
\definecolor{blue}{rgb}{0, 0.2, 0.8}
\definecolor{orange}{rgb}{0.6, 0.6, 0}
\definecolor{red}{rgb}{0.8, 0.2, 0.2}
\definecolor{magenta}{rgb}{0.5, 0.0, 1.0}
\definecolor{black}{rgb}{0.0, 0.0, 0.0}
\definecolor{cyan}{rgb}{0, 0.65, 0.65}

\newif\ifdraft
\draftfalse

\ifdraft
\newcommand{\dlc}[1]{{\color{blue}\textbf{DL:} #1}}
\newcommand{\dcc}[1]{{\color{red}\textbf{DC:} #1}}
\newcommand{\kac}[1]{{\color{orange}\textbf{KA:} #1}}
\newcommand{\rwc}[1]{{\color{green}\textbf{RW:} #1}}

\else
\newcommand{\dlc}[1]{}
\newcommand{\dcc}[1]{}
\newcommand{\kac}[1]{}
\newcommand{\rwc}[1]{}

\fi

\newcommand{\bp}{{\bf p}}

\newcommand{\Loss}{\mathcal{L}}
\newcommand{\mm}{\mathcal{M}}

\newcommand{\ms}{\mathcal{S}}
\newcommand{\pp}{\mathcal{P}}
\newcommand{\bbe}{\mathbb{E}}
\def \figures {./}

\begin{document}

\title{Learning Character-Agnostic Motion for Motion Retargeting in 2D}

\author{Kfir Aberman}
\affiliation{%
  \institution{Tel-Aviv University,}
  \institution{AICFVE Beijing Film Academy}
  }

\author{Rundi Wu}
\affiliation{%
  \institution{Peking University}
}  

\author{Dani Lischinski}
\affiliation{%
  \institution{Shandong University, Hebrew University of Jerusalem}
  }
  
\author{Baoquan Chen}
  \authornote{Corresponding author}
\affiliation{%
  \institution{Peking University}
}
  
  \author{Daniel Cohen-Or}
\affiliation{%
  \institution{Tel-Aviv University}
}

\renewcommand\shortauthors{Aberman, K. et al}

\begin{abstract}

Analyzing human motion is a challenging task with a wide variety of applications in computer vision and in graphics.
One such application, of particular importance in computer animation, is the retargeting of motion from one performer to another.
While humans move in three dimensions, the vast majority of human motions are captured using video, requiring 2D-to-3D pose and camera recovery, before existing retargeting approaches may be applied.
In this paper, we present a new method for retargeting video-captured motion between different human performers, without the need to explicitly reconstruct 3D poses and/or camera parameters.	

In order to achieve our goal, we learn to extract, directly from a video, a high-level latent motion representation, which is invariant to the skeleton geometry and the camera view. Our key idea is to train a deep neural network to decompose temporal sequences of 2D poses into three components: motion, skeleton, and camera view-angle.
Having extracted such a representation, we are able to re-combine motion with novel skeletons and camera views, and decode a retargeted temporal sequence, which we compare to a ground truth from a synthetic dataset.

We demonstrate that our framework can be used to robustly extract human motion from videos, bypassing 3D reconstruction, and outperforming existing retargeting methods, when applied to videos in-the-wild. It also enables additional applications, such as performance cloning, video-driven cartoons, and motion retrieval.

Webpage (code and data): \href{https://motionretargeting2d.github.io/}{https://motionretargeting2d.github.io/}

\end{abstract}

%

%
%




\maketitle

\section{Introduction}
\label{sec:intro}

\begin{figure}
\centering
\includegraphics[width=\linewidth]{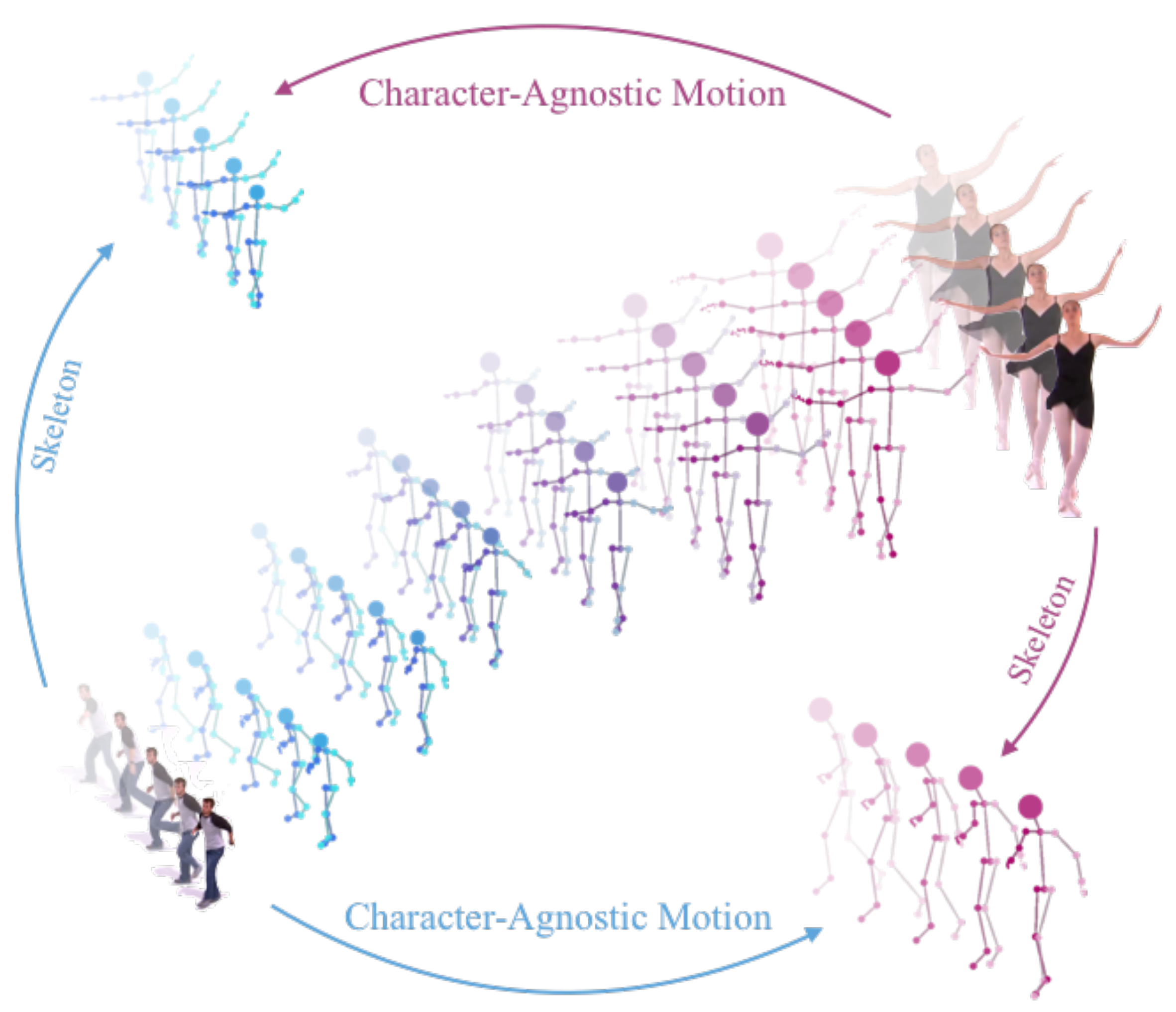} 
\caption{Given two videos of different performers, our approach enables to extract character-agnostic motion from each video, and transfer it to a new skeleton and view angle (top-left and bottom-right), directly in 2D. In addition, separate latent representations for motion, skeleton, and view-angle are extracted, enabling control and interpolation of these parameters.}
\label{fig:teaser}
\end{figure}

Understanding and synthesizing human motion has been a central research topic in computer animation.
Motion is inherently a 4D entity, commonly represented using a low-level encoding: as a temporal sequence of poses, specified as a set of joint positions and/or angles.
Such a representation strongly depends on the skeleton and its geometric properties, such as the lengths of the limbs and their proportions.
Thus, the same motion performed by two individuals with different skeletons might have significantly different representations.
One might even argue that character-agnostic motion is a slippery and elusive notion, which is not completely well-defined.

In this work, we address the challenging problem of retargeting the video-captured motion of one human performer to another.
In a nutshell, our approach is to extract an abstract, character- and camera-agnostic, latent representation of human motion directly from ordinary video. The extracted motion may then be applied to other, possibly very different, skeletons, and/or shown from new viewpoints.

The challenges that we face are twofold: First, the abstract motion representation that we seek is new and unknown, and thus we do not have the benefit of supervision. Second, working on video introduces an additional obstacle, as the joint trajectories are observed in 2D, and thus are not only character-specific, but also view-dependent, suffering from ambiguities and occlusions.

\begin{figure}
	\centering
	\includegraphics[width=\linewidth]{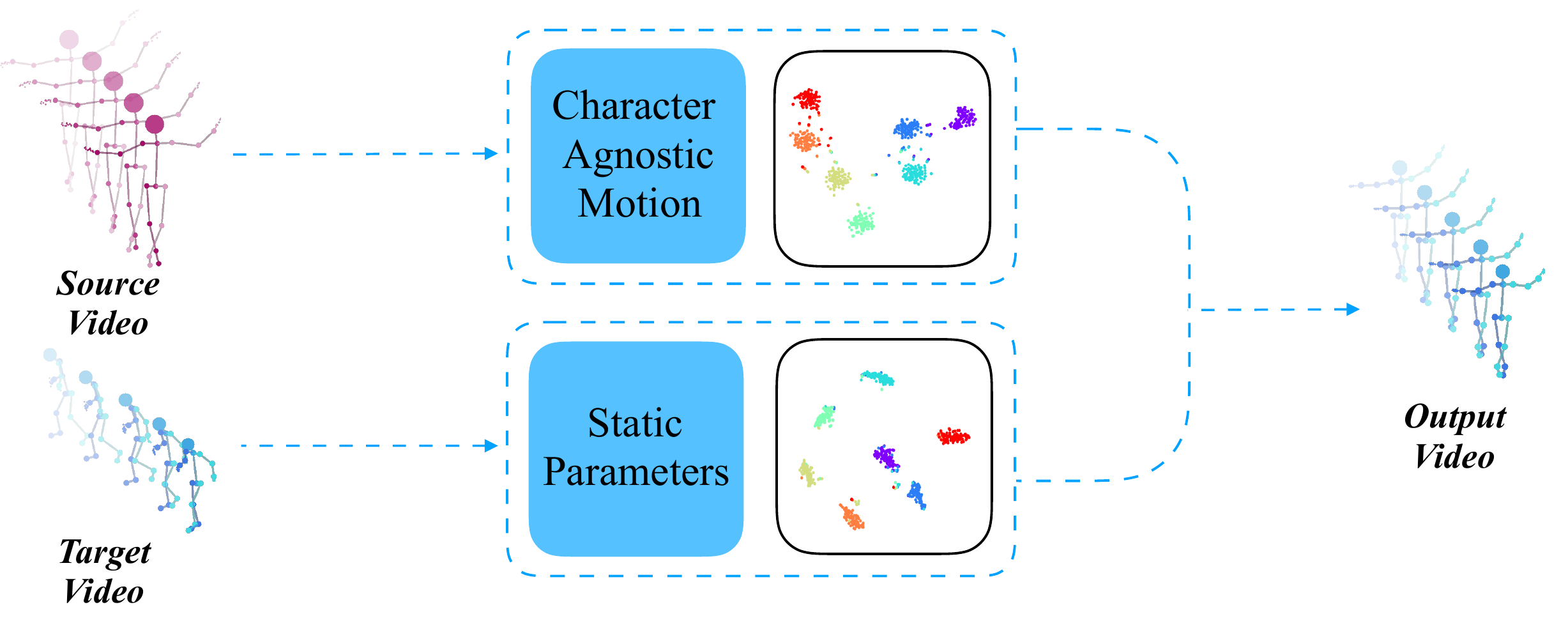} 
	(a)
		\includegraphics[width=\linewidth]{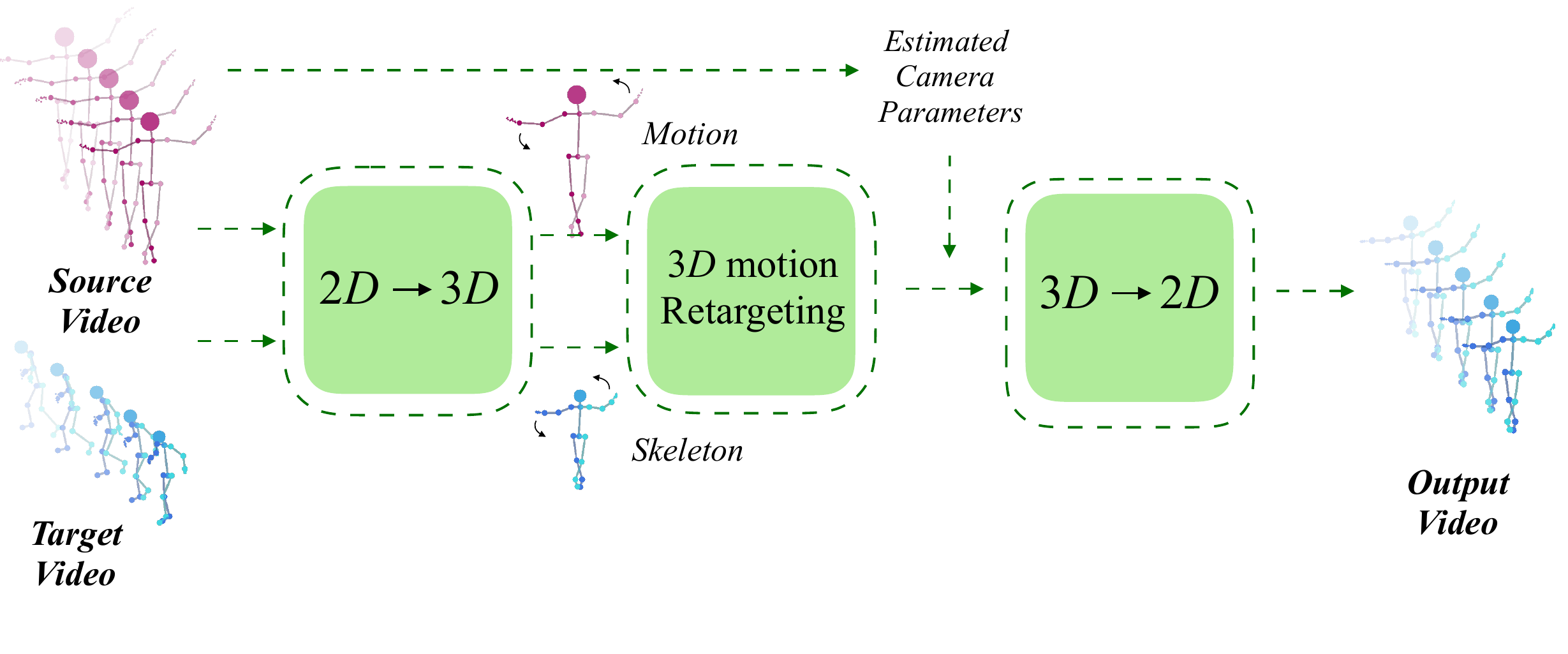} 
	(b)
	\caption{Our network learns a dynamic character-agnostic latent motion representation, along with static latent components. This enables motion retargeting directly in the 2D domain (a), bypassing the need for ambiguous 2D-to-3D pose and camera parameters estimation (b).}
	\label{fig:bypass_3d}
\end{figure}

Our motivation for learning directly from 2D videos stems from the fact that the vast majority of existing depictions of human motion are captured in this way. 
Furthermore, despite impressive recent progress in 3D human pose recovery from video, enabled by recent deep learning machinery, this is still an error-prone process, which we bypass by working directly in 2D, as illustrated in Figure \ref{fig:bypass_3d}.

The key idea behind our approach is to train a deep neural network to perform 2D motion retargeting, which learns, in the process, to extract three separate latent components:
(i) a dynamic component, which is a skeleton-independent and view-independent encoding of the motion, (ii) a static component, which encodes the performer's skeleton, and (iii) a component that encodes the view-angle. 
The last component may be either static or dynamic (depending on whether the camera is stationary or not), but in this work we assume it is static.
Once extracted, these latent components are recombined to yield new motions, allowing a loss to be computed and optimized. As pointed out earlier, the same motion performed by different individuals cannot be expected to be truly identical in the corresponding latent space. Thus, in practice, we implicitly learn to cluster motions in the dynamic latent space, where each cluster consists of similar motions performed by different individuals.

In practice, our architecture consists of three encoders that decompose an input sequence of 2D joint positions into the aforementioned latent spaces, and a decoder that reconstructs a sequence from such components. Since motion sequences may differ in length, our encoders are designed such that the resulting latent motion representation is duration-dependent, while the other two attributes are encoded into a duration-independent latent space.

We train the network to decompose 2D projections of synthetic 3D data into these three attributes, which are then shuffled and re-composed to form new combinations. Since the training data is synthetic, the ground truth can be generated by motion retargeting in 3D, while respecting physical constraints. Specifically, we use Adobe Mixamo \cite{mixamo} to obtain sequences of poses of different 3D characters, with different skeletal properties, which perform the same motion and follow kinematic constraints.

During training we use augmentation and add artificial noise to simulate occlusions and errors that one might encounter in real videos. We demonstrate that at test time our network can be successfully applied to videos in-the-wild, with better accuracy than existing alternatives.
In particular, we show that on such videos we outperform motion retargeting methods that operate in 3D, mainly because of their dependence on reliable 3D pose estimation from video (see Figure~\ref{fig:bypass_3d}).
We also show that the learned latent spaces are continuous, enabling independent interpolation of motion, skeletons, and views between pairs of sequences, as illustrated in Figure~\ref{fig:teaser}.
In summary, our results demonstrate that deep networks can constitute a better solution for specific sub-tasks, which do not strictly require a full 3D reconstruction.




\section{Related Work}
\label{sec:related}

\subsubsection{Motion Representation}
M\"{u}ller \etal~\shortcite{muller2009efficient} propose to represent motion as an explicit matrix that captures the consistent and variable aspects of learned motion classes. Unknown motion inputs are segmented and annotated by locally comparing them with the available motion templates. Bernard \etal~\shortcite{bernard2013motionexplorer} developed MotionExplorer, an exploratory search system that clusters and displays motions as a hierarchical tree structure. Their method combines a number of visualization techniques to support user overview and exploration. The authors apply the divisive hierarchical clustering algorithm to the low-level pose features, and train a self-organizing map (SOM) on all feature vectors in order to arrange them in a topology preserving grid.

Similarly, Wu \etal~\shortcite{wu2009indexing}, and later Hu \etal~\shortcite{hu2010motion}, cluster motion on hierarchically structured body segments, and measure the temporal similarity of each partition using SOM, which is computationally expensive. Chen \etal~\shortcite{chen2015scalable} used hierarchical affinity propagation (HAP) to perform data abstraction on low-level pose features to generate multiple layers of data aggregations. Bernard \etal~\shortcite{bernard2017visual} present a visual-interactive approach for the semi-supervised labeling of human motion capture data; users assign labels to the data which can subsequently be used to represent the multivariate time series as sequences of motion classes. Recently, Aristidou \etal~\shortcite{Aristidou:2018:TOG} mapped motion words from 3D captured data into a latent space. In all of these approaches, the analysis is performed on full 3D motion data, and synthesis is not addressed.

Tulyakov \etal~\shortcite{tulyakov2017mocogan} designed a GAN that is fed by two noise vectors, a time dependent and a time independent one, in order to generate video frames that can be separately controlled by motion and content.
While this approach also generates motion by combining static and dynamic components, they do not explore the decomposition of a given motion to such components. 

Inverse graphics networks \cite{kulkarni2015deep}, learn an interpretable representation of images by decomposing them into shape, pose and lighting codes.
Peng \etal~\shortcite{peng2017reconstruction} disentangle face appearance from its pose, by learning a pose-invariant feature representation. Ma \etal~\shortcite{ma2018disentangled} disentangle and encode background, foreground, and pose from still human images into embedding features, which are then combined to re-compose the input image. In contrast, we learn to disentangle motion data directly from a video, using synthetic data as ground truth to compare with the re-composed motion.

Holden \etal~\shortcite{holden2015learning} used an auto-encoder to learn the motion manifold of uni-sized 3D characters from motion capture data, and later on used this representation to synthesize character movements based on high level parameters \cite{holden2016deep}. Since they use a normalized skeleton, their approach is not applicable to motion retargeting between different skeletons, in contrast to our approach, which extracts a skeleton-specific static latent feature.



\subsubsection{Motion Retargeting}
Our system extracts motion from videos of humans by performing a supervised 2D motion retargeting. However, since most of the existing motion retargeting methods operate in 3D, we next survey a few works in that domain.

Gleicher \etal~\shortcite{gleicher1998retargetting} first formulated motion retargeting as a spacetime optimization problem with kinematic constraints, which is solved for the entire motion sequence. Lee and Shin \shortcite{lee1999hierarchical} proposed a decomposition approach that first solves the IK problem for each frame to satisfy the constraints and then fits multilevel B-spline curves to achieve smooth results. Tak and Ko \shortcite{tak2005physically} further added dynamics constraints to perform sequential filtering to render physically plausible motions. Choi and Ko \shortcite{choi2000online} propose an online retargeting method by solving per-frame IK that computes the change in joint angles corresponding to the change in end-effector positions, while imposing motion similarity as a secondary task.

While the aforementioned approaches require iterative optimization with hand-designed kinematic constraints for particular motions, our method learns to produce proper and smooth changes of joint positions in a single feed-forward inference pass through our network, and is able to generalize to unseen characters and novel motions. 

The idea of solving approximate IK can be traced back to the early blending-based methods \cite{kovar2004automated,rose2001artist}. A target skeleton may be viewed as a new style. Our method can be applied to motion style transfer, which has been a popular research area in computer animation \cite{hsu2005style, min2010synthesis, xia2015realtime}. Recently, Villegas \etal~\shortcite{villegas2018neural} proposed a recurrent neural network architecture with a Forward Kinematics layer and cycle consistency based adversarial training objective for unsupervised motion retargeting.

All of the works mentioned above perform the motion retargeting in 3D, in contrast to our approach, which leverages the abilities of deep networks to learn mappings between 2D input and output, thereby bypassing the need for 3D human pose and camera pose recovery from 2D data.

Peng \etal~\shortcite{peng2018sfv} propose a method that enables physically simulated characters to learn skills from videos (SFV), based on deep pose estimation and deep reinforcement learning. Although they learn from video, the learned skills are applied to 3D characters, and their results critically depend on the accuracy of 3D pose estimation.

\section{Motion Learning Framework}
\label{sec:network}


At the crux of our approach lies a multi-encoder/single-decoder neural network trained to decompose and recompose temporal sequences of 2D joint positions. The network encodes the input samples into three separate feature spaces: (i) a dynamic, skeleton- and view-independent, motion representation, (ii) a static skeleton-dependent feature, and (iii) a view-dependent feature. The latent representation of the motion is duration-dependent, while the two latter features reside in a duration-independent latent space.

To train such a network, we leverage a synthetic dataset that comprises temporal sequences of 2D poses of different characters, each performing a set of similar motions. The learning is indirectly-supervised, namely, no ground truth exists for the desired motion representation; however, we do have multiple samples of each motion, as performed by the different characters, and these motions can be projected to 2D, from arbitrary view angles. Thus, by forcing the network to decompose the provided motion samples, followed by shuffling the components and re-composing new ones, the training ensures that each of the extracted components indeed encodes the intended information.
\begin{figure}
	\includegraphics[width=\linewidth]{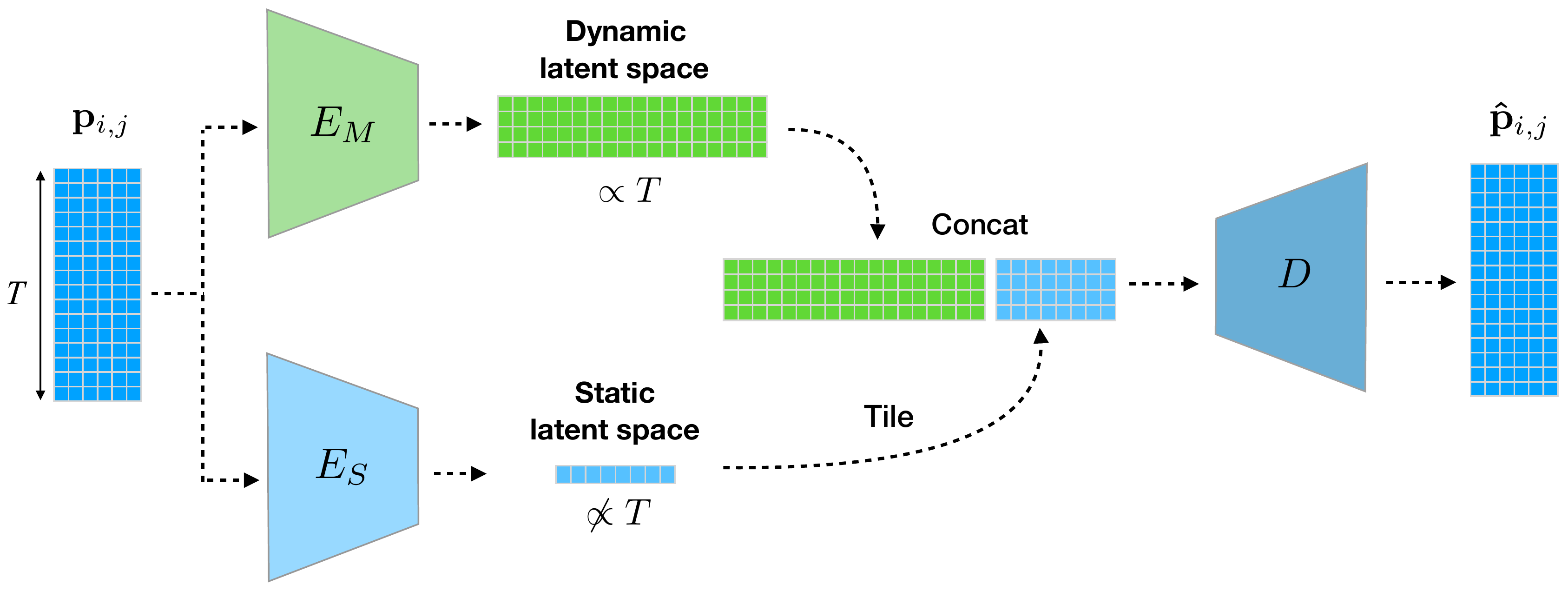} \\
(a)\\
\includegraphics[width=\linewidth]{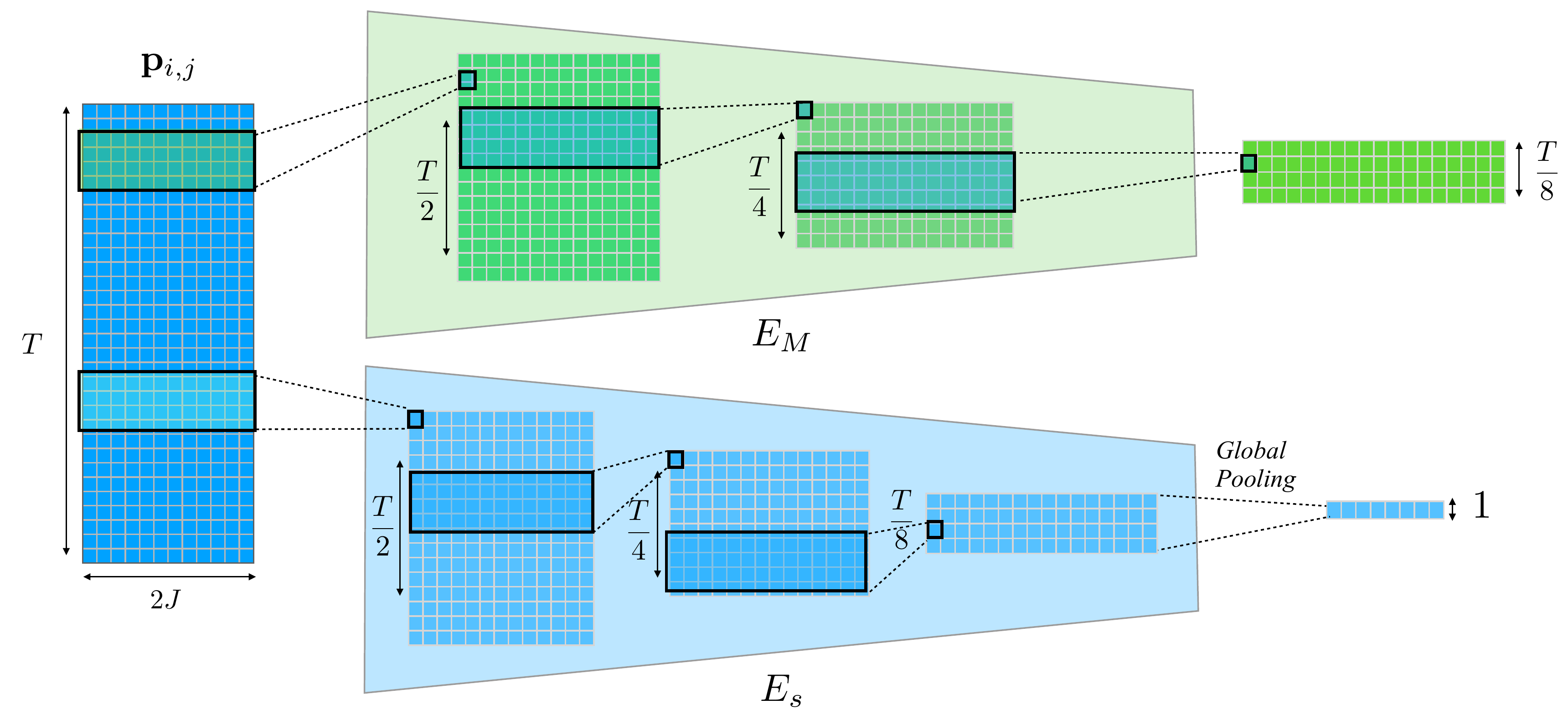} 
(b)\\
	\caption{Our framework encodes dynamic (duration-dependent) and static (duration-independent) features into separate latent spaces, using two encoders, $E_M$ and $E_S$. In order to decode the sequence with the decoder $D$, the static latent feature is tiled along the temporal axis, and concatenated to the motion latent feature along the channel axis. (b) We use one-dimensional convolution layers with stride 2, over the temporal dimension, letting $E_M$ generate a latent motion whose size depends on the duration of the input sample, while in $E_S$ a global pooling layer is employed along the temporal axis to collapse it, resulting in a latent vector of a fixed size.}
	\label{fig:high_level_scheme}
\end{figure}

\subsection{Network Architecture}
\label{subsec:network}
For clarity of exposition, in the following section we regard the view and the skeleton as a single static attribute, and describe our framework using two attributes, dynamic and static, extracted by two encoders. The derivation is easily extended to three attributes, extracted by three encoders.

Let $\mm$ and $\ms$ denote the set of different motions and the set of different static attributes, respectively. 
Let $\bp_{i,j}\in\mathbb{R}^{T\times 2J}$ be a data sample that can be described by two attributes, dynamic ($i\in\mm$) and static ($j\in\ms$), where $T$ is the temporal length of the motion, and $J$ is the number of joints (each joint is specified by its 2D coordinates).

A high level diagram of our approach is shown in Figure~\ref{fig:high_level_scheme}(a). Each data sample is encoded, in parallel, by two encoders, $E_M$ and $E_S$, whose output is then concatenated and fed into a decoder $D$. Our goal is to train the network to decompose the motion sample into two separate latent codes, one capturing the dynamic aspects of the motion, and another capturing the static aspects. In order to encourage this, the $E_M$ encoder is designed to preserve the temporal information, using one-dimensional convolution layers, with strides, over the temporal dimension, and generate a latent motion whose size depends on the duration of the input sample (downsampled by a fixed factor). In contrast, the $E_S$ encoder, employs global pooling to collapse the temporal axis, resulting in a latent vector of a fixed size, independent of the input sequence length, as illustrated in Figure~\ref{fig:high_level_scheme}(b). Thus, the network, which is trained on various sequence lengths, learns to separate dynamic-static attributes.
The two latent features are combined, before being fed into the decoder $D$, by tiling (replicating) the static, fixed-length, features along the temporal axis and then concatenating the two parts the along the channel axis (see Figure~\ref{fig:high_level_scheme}(a)).

\subsection{Decomposition and Re-composition}
\label{subsec:decompose}
Although the structure of the network explicitly separates duration-dependent dynamic features from static ones, this in itself cannot ensure that the dynamic feature necessarily encodes skeleton/view-agnostic motion, since there are arguably many possible dynamic-static decompositions. In order to force the network to perform the desired decomposition, we train it with our synthetic data, which demonstrates what similar motions look like when applied to different characters and projected onto different views.
The key idea is to require that various combination of latent motions and static parameters can be used to reconstruct the corresponding ground truth samples. Formally, given two data samples, $\bp_{i,j}, \bp_{k,l} \in \mathbb{R}^{T\times 2J}$, where $i,k\in\mm$ and $j,l\in\ms$, we ideally want that
\begin{equation}
\forall_{i,k\in\mm,\; j,l\in \ms} \;\;\bp_{i,l} \approx D\left(E_M\!\left( \bp_{i,j} \right),  E_S\!\left( \bp_{k,l} \right) \right). 
\label{eq:cross_cond}
\end{equation}
The above requirement encourages the encoders $E_M$ and $E_S$ to map input samples, with similar attributes, into tightly clustered groups in the corresponding latent spaces; these clusters may then be mapped back into various samples that share the same attribute. The relationship between the condition in \eqref{eq:cross_cond} to clustering is demonstrated via experiments in the Section~\ref{sec:results}. In Section \ref{subsec:loss} we explain how this re-composition concept is applied, during training. 

\begin{figure}
	\includegraphics[width=\linewidth]{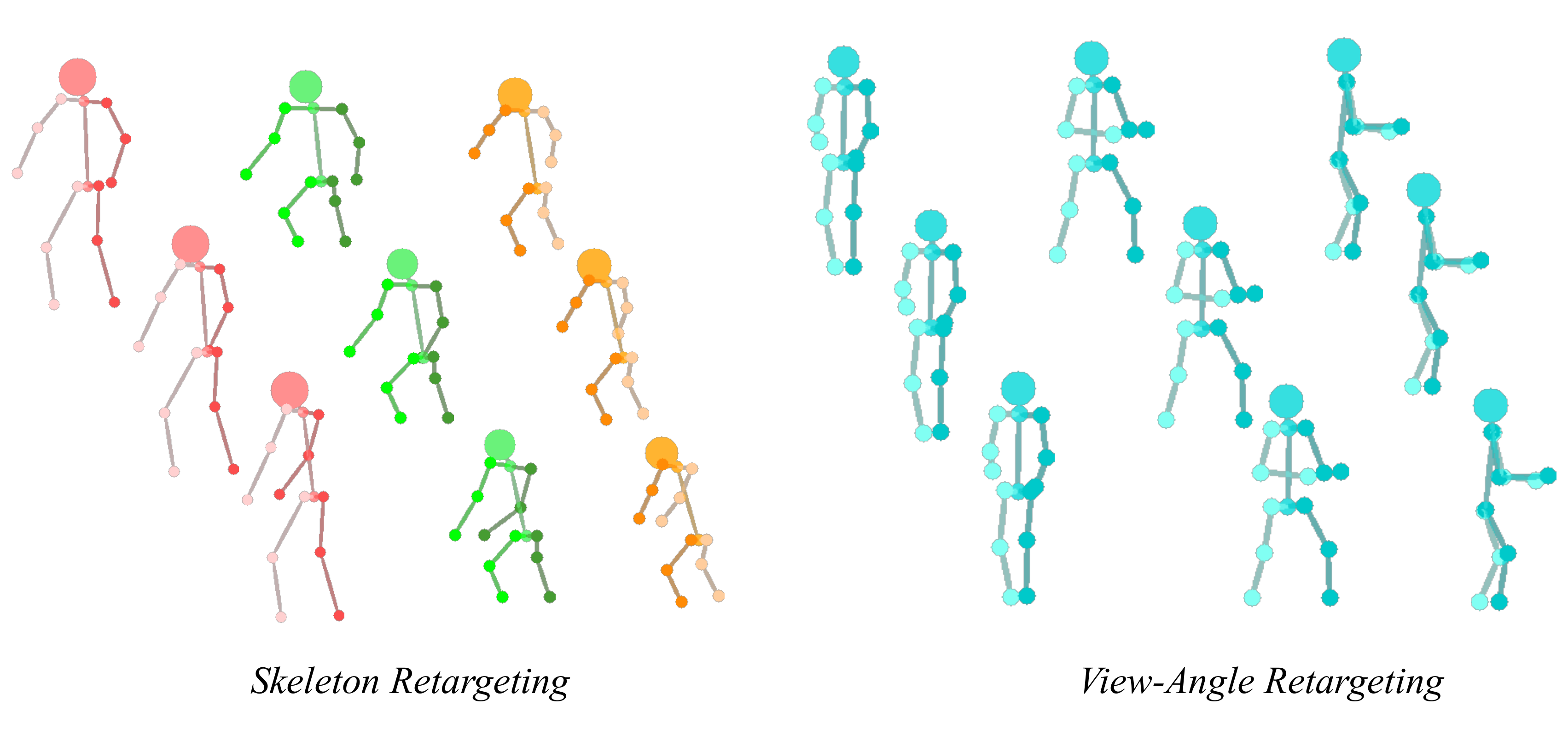} 
	 \\ (a)
	\includegraphics[width=\linewidth]{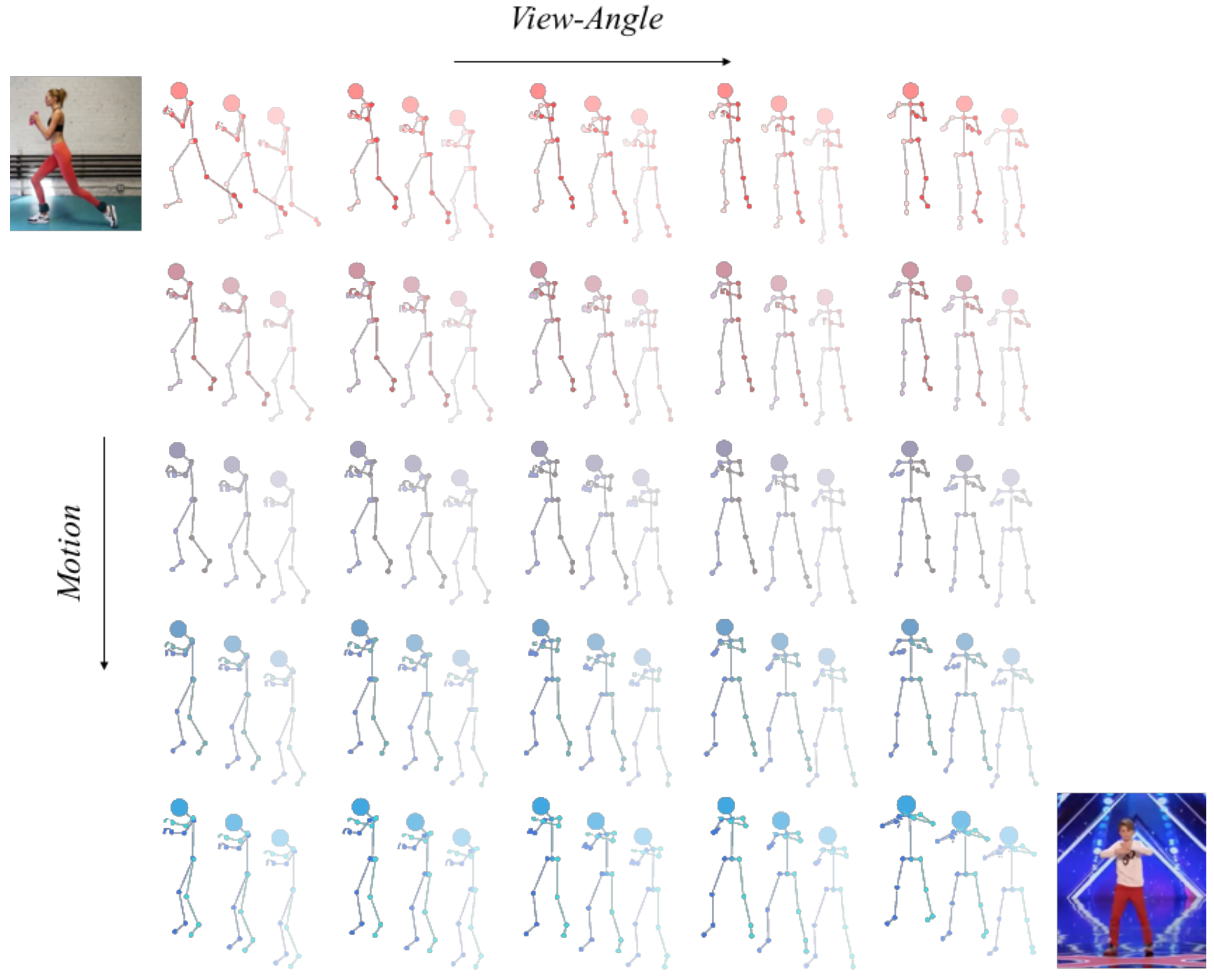} 
	 \\ (b)
	\caption{Retargeting and interpolation made possible by decomposing motions into three separate latent spaces. (a) Retargeting of similar motion to various skeletons (left) and different view-angles (right). (b) Interpolation of view-angle (horizontal axis) and motion (vertical axis).}
	\label{fig:results}
\end{figure}

The separate dynamic and static latent spaces learned by our network enable a variety of manipulations, such as retargeting the motion to a different skeleton, or to a different view, as well as continuously interpolating skeletons, views, and motions, as demonstrated in Figure \ref{fig:results}.

\subsection{Training and Loss}
\label{subsec:loss}
To train our network, we use a loss function consisting of three components: cross reconstruction loss, triplet loss, and foot velocity loss. These components are described in more detail below.

\begin{figure}
	\includegraphics[width=\linewidth]{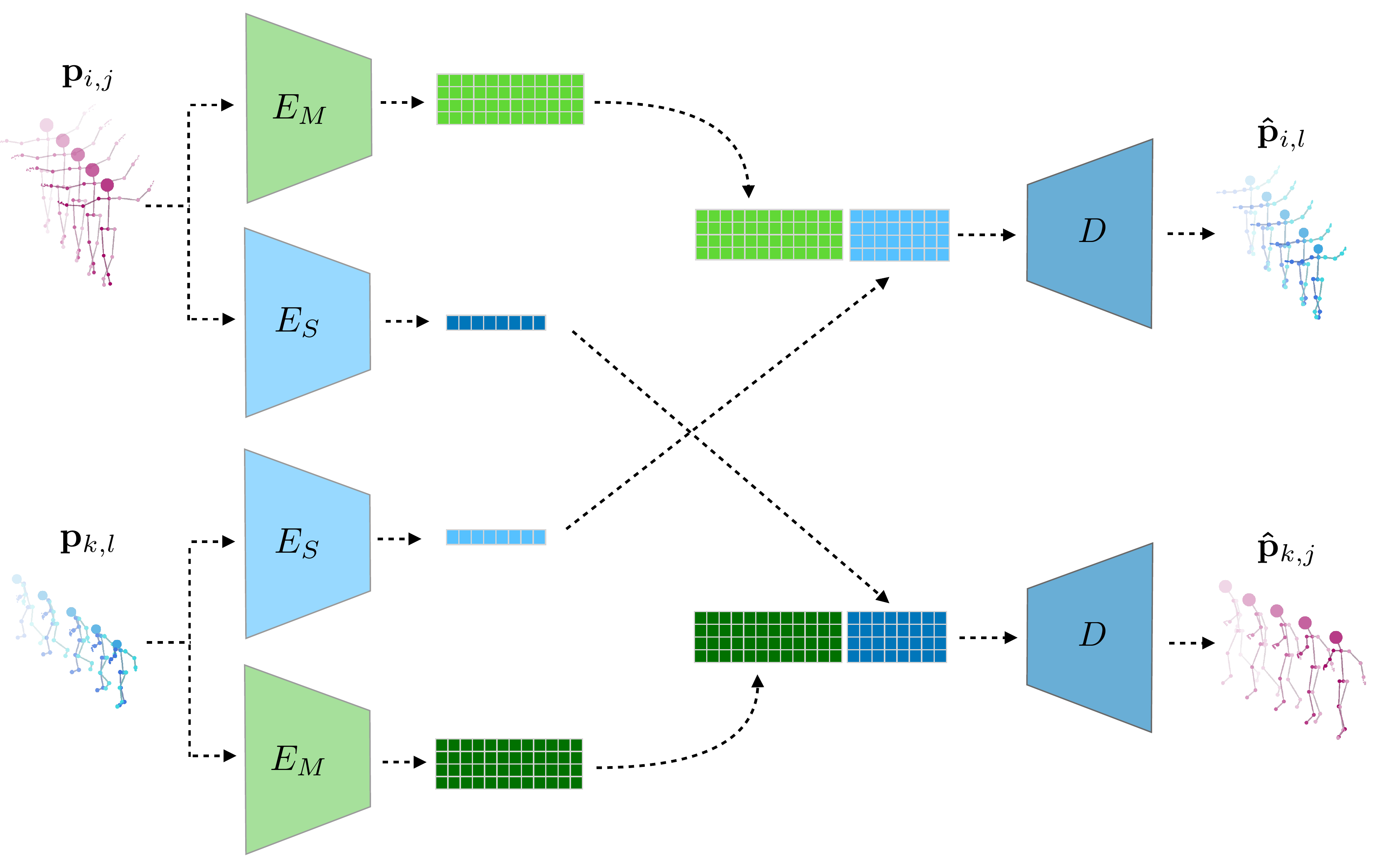} 
	\caption{Cross reconstruction loss: the static features extracted from two input samples are swapped, and recombined with the dynamic features.}
	\label{fig:cross_loss}
\end{figure}

\paragraph{Cross Reconstruction Loss}
In order to achieve the implicit separation via the condition in \eqref{eq:cross_cond} we train our network to reconstruct cross compositions of various pairs, as illustrated in Figure~\ref{fig:cross_loss}.

In practice, in each iteration, we randomly draw a pair of samples from the training dataset $\pp$, decompose them by the encoders and re-compose new combinations using the decoder. Since the ground truth exists in the dataset, we can explicitly require:
\begin{eqnarray}
\Loss_{\text{cross}} & = & \bbe_{\bp_{i,j},\bp_{k,l}\sim \pp\times\pp}\left[ \Vert D(E_M(\bp_{i,j}),E_S(\bp_{k,l}))-\bp_{i,l}\Vert^2 \right] \\\nonumber
&+& \bbe_{\bp_{i,j},\bp_{k,l}\sim \pp\times\pp}\left[ \Vert D(E_M(\bp_{k,l}),E_S(\bp_{i,j}))-\bp_{k,j}\Vert^2\right].
\label{eq:cross_loss}
\end{eqnarray}
The number of drawn pairs in each epoch is equal to the number of samples in the training data.

In addition to the cross reconstruction requirement, in every iteration we also require that the network reconstructs each of the original input samples, which can be formulated as a standard autoencoder reconstruction loss:
\begin{equation}
\Loss_{\text{rec}} = \bbe_{\bp_{i,j}\sim \pp}\left[ \Vert D(E_M(\bp_{i,j}),E_S(\bp_{i,j}))-\bp_{i,j}\Vert^2 \right].
\end{equation}
The above losses are combined together to $\Loss_{\text{cross_rec}} = \Loss_{\text{rec}} + \Loss_{\text{cross}}$.

\paragraph{Triplet loss}
Cross reconstruction loss by itself ensures that latent vectors of similar motions are decoded into a sequence that exhibits this motion. However, since there is no explicit requirement for separation between the different attributes, the latent space of one attribute may still contain information about the other, as demonstrated in Section~\ref{sec:ablation}. 

In order to enhance the separation and explicitly encourage samples with similar motion to be mapped tightly together, we use the techinque of Aristidou \etal~\shortcite{Aristidou:2018:TOG}, to map samples with similar motions into the same area, and directly apply a triplet loss on the motion latent space:
\begin{align}
\label{eq:triplet}
\Loss_{\text{trip_M}} & = & \bbe_{\bp_{i,j},\bp_{i,l},\bp_{k,l}\sim \pp}
[ &\Vert E_M(\bp_{i,l}) - E_M(\bp_{i,j})\Vert  - &\\\nonumber
&  & &\Vert E_M(\bp_{i,l})- E_M(\bp_{k,l})\Vert + \alpha ]_{+}, 
\end{align}
where $i\neq k$, and $\alpha=0.3$ is our margin. This loss takes care to place the projection of two samples that share the same motion at a distance that is smaller (at least by $\alpha$) than the distance between two samples with different motions. In practice, in every iteration, we use the drawn pair and the corresponding cross ground truth to pick two triplets, where each contains a pair that shares the same motion.
The same triplet concept is applied to the latent space of the static parameters $\Loss_{\text{trip_S}}$, which is defined as in \eqref{eq:triplet}, with $E_S$ instead of $E_M$.  Summing the two parts leads to a total triplet loss of $\Loss_{\text{trip}} = \Loss_{\text{trip_M}} + \Loss_{\text{trip_S}}$.

Our experiments show that this additional constraint, not only leads to a better disentanglement but also to a better retargeting (Section \ref{sec:ablation}). An alternative constraint would be to directly require that two samples corresponding the the same motion should be mapped into the same point in the latent space. However, our experiments indicate that such a requirement is too strict, and results in degraded retargeting performance.
In addition, it should be noted that using a simple (non-cross) reconstruction loss along with the triplet loss proves insufficient for retargeting and transfer, as shown in our ablation study (Section \ref{sec:ablation}).

\paragraph{Foot velocity loss}
Using only a reconstruction loss, our experiments show that end effectors, such as hands and feet exhibit larger errors, which gives rise to the well-known foot skating phenomenon. The reason is that, even though the network is trained to reconstruct the original poses, it will prefer to put its efforts on strategic central joint positions that have a greater influence on the rest of the body. Thus, we explicitly constrain the global positions of the end-effectors ($\mathcal{J}_{\text{end}}$), which is essential for fixing foot sliding artifacts or guiding the hand of the character to grasp objects, by
\begin{equation}
\Loss_{\text{foot}} = \bbe_{\bp_{i,j}\sim \pp}\sum_{n\in \mathcal{J}_{\text{end}}}\Vert V_{\text{global}}(\hat{\bp}_{ij}) + V_{\text{joint}_n}(\hat{\bp}_{ij})  - V_{\text{orig}_n}(\bp_{ij}) \Vert^2,
\label{eq:foot_loss}
\end{equation}
where $V_{\text{global}}$ and $V_{\text{joint}_n}$ extract the global and local ($n$th joint) velocities from the reconstructed output $\hat{\bp}_{ij}$, respectively, and map them back to the image units, and $V_{\text{orig}_n}$ returns the original global velocity of the $n$th joint from the ground truth, $\bp_{ij}$. The contribution of $\Loss_{\text{foot}}$ to the mitigation of the foot skating phenomena is demonstrated in the supplementary video.

Summing the three terms, we obtain our total loss:
\begin{equation}
\Loss = \Loss_{\text{cross_rec}}+\lambda_1\Loss_{\text{trip}} + \lambda_2\Loss_{\text{foot}},
\label{eq:total_loss}
\end{equation}
where in all of our experiments $\lambda_1 = 0.1$ and $\lambda_2 = 0.5$.

Following training, the weights of the learned filters exhibit strong inter-joint correlations.
For example, it appears that most of the joint filters in the view-angle encoder learn to observe the hips and the shoulders, whose width on the image plane is more indicative of the view angle than the limbs.

\subsection{Motion Dataset}
\label{subsec:dataset}

We constructed our 2D motion dataset using the Mixamo \cite{mixamo} 3D animation collection, which contains approximately 2400 unique motion sequences, including elementary actions (jumping, kicking, walking, etc.), and various dancing moves (samba, hip-hop, etc.). Each of these motions may be applied to 71 distinct characters, which share a human skeleton topology, but may differ in their body shape and proportions. The motions are automatically adapted to the different characters using the 3D motion retargeting algoritm Human-IK of AutoDesk \cite{montgomery2012tradigital}.

\begin{figure}
\includegraphics[width=\linewidth]{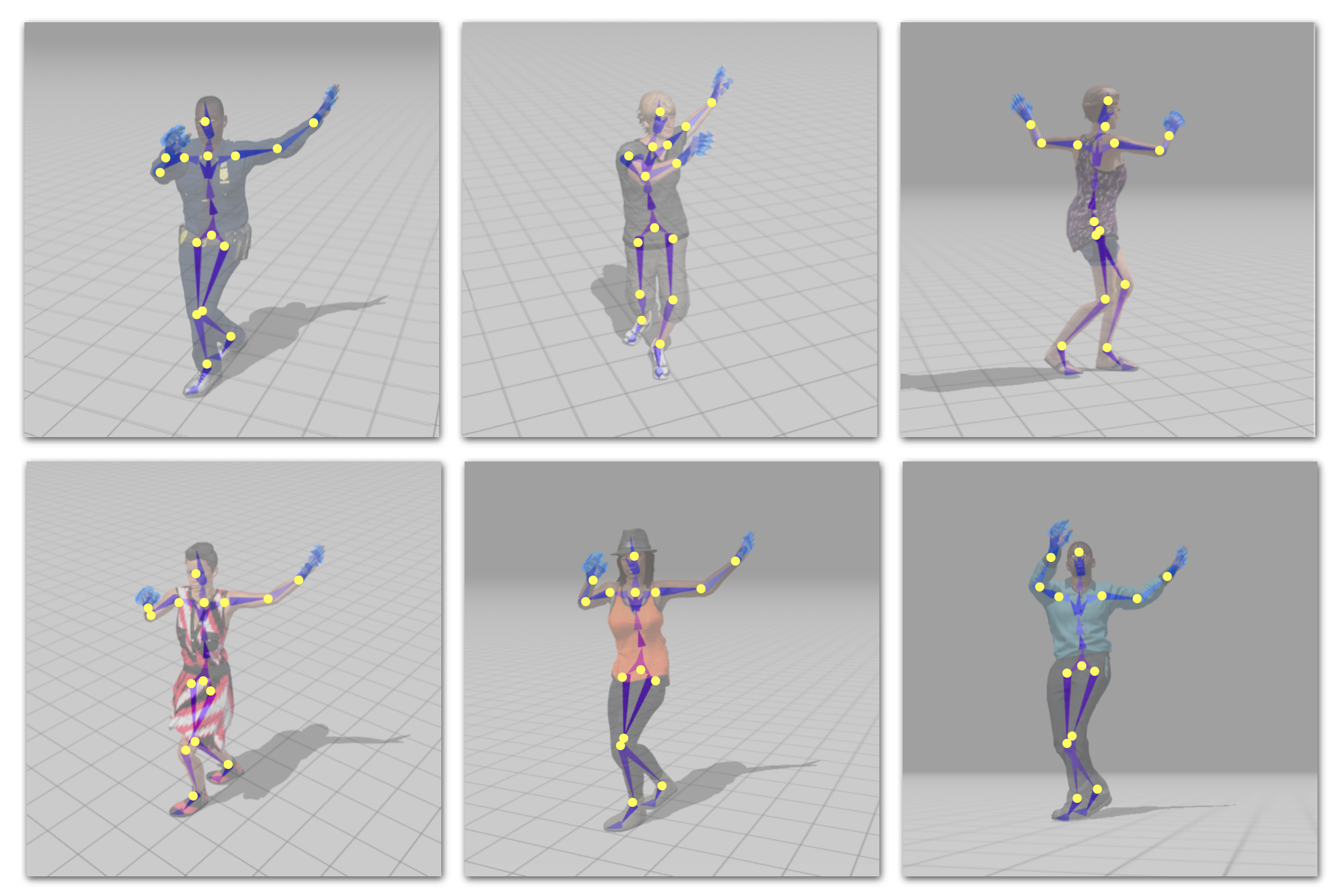} 
\caption{We use Mixamo  \cite{mixamo} to construct our 2D motion dataset. A variety of 3D characters, which differ in their skeleton geometry, each perform a set of similar motions. The dataset is constructed by projecting the positions of selected joints (shown above as yellow dots) into 2D, using a variety of view angles.}
\label{fig:mixamo}
\end{figure}

In practice, we generate our data samples by projecting the 3D joint positions of characters, performing similar motions, into different camera view angles, as illustrated in Figure~\ref{fig:mixamo}. As a result, we obtain a rich labeled dataset, consisting of over 500,000 samples, which demonstrates how skeletons of different characters, that perform similar motions, appear from different views. 

Formally, for a given set of motions ($\mathcal{M}$) and characters ($\mathcal{C}$), let $f^{(t)}(i,k)\in\mathbb{R}^{3\times J}$ denote a matrix that contains the $J$ 3D joint positions of character $k\in\mathcal{C}$ at time $t$, while performing the motion $i\in\mathcal{M}$. $f$ may be thought of as the query function that extracts the appropriate pose from the dataset. Then, the projections to various view angles ($\mathcal{V}$) are performed using the weak-perspective camera model which consists of a rotation matrix $R_v\in\mathbb{R}^{3\times 3}$ in axis-angle representation, translation $b\in\mathbb{R}^2$, and scale $s\in\mathbb{R}$, yielding
\begin{equation}
p_{i,k,v}^{(t)} = s\Pi (R_vf^{(t)}(i,k)) + b,
\label{eq:weak_projection}
\end{equation}
where $\Pi$ is an orthographic projection. The rotation $R_v$ is defined in the character's temporal average coordinate system that is computed during its motion period, where the forward direction (Z-axis) in each time step is computed by the cross product of the vertical axis (Y-axis) and the average of the vector across the left and right shoulders with the vector across the hips. In the data generation step the scaling and translation are taken as constants,  $b=(0,0)$, $s=1$, and will be augmented during training. 

We partition the frames into temporal windows of $T=64$ to construct our dataset samples, $\bp_{i,k,v} \in \mathbb{R}^{T\times 2J}$, where $i$, $k$ and $v$ indicate the indices of the motion ($i\in\mm$), skeleton ($k\in\mathcal{C}$), and camera view angle ($v\in\mathcal{V}$), respectively.
Since we want to apply the system to real videos at test time, we selected $J=17$ joints that appear both in the 3D skeletons in the dataset and in the method of Cao \etal~\shortcite{cao2016realtime} (BODY_25 representation), which is used for 2D pose estimation. The joints, which constitute a basic skeleton (head, neck, shoulders, hips, knees, ankles, toes, heels, elbows, and wrists), are shown as yellow dots in Figure~\ref{fig:mixamo}. We further use the method of Simon et al.~\shortcite{simon2017hand} to detect 3 joints per finger, yielding 30 additional joints.


\paragraph{Preprocessing}
To normalize the data we first globally subtract the root position from all joint locations in every frame, then locally (per joint) subtract the mean joint position and divide by the standard deviation (averaged over the entire dataset). These operations are invertible, so the original sequence can be restored after the reconstruction. The normalized representation does not contain global information, thus, we omit the root position (which is permanently zero) and append the per-frame global velocity, in the image plane (XZ), to the input representation. The velocities can be integrated over time to recover the global translation of the character.

\subsection{Implementation Details}
\begin{figure}
	\centering
	\includegraphics[width=\linewidth]{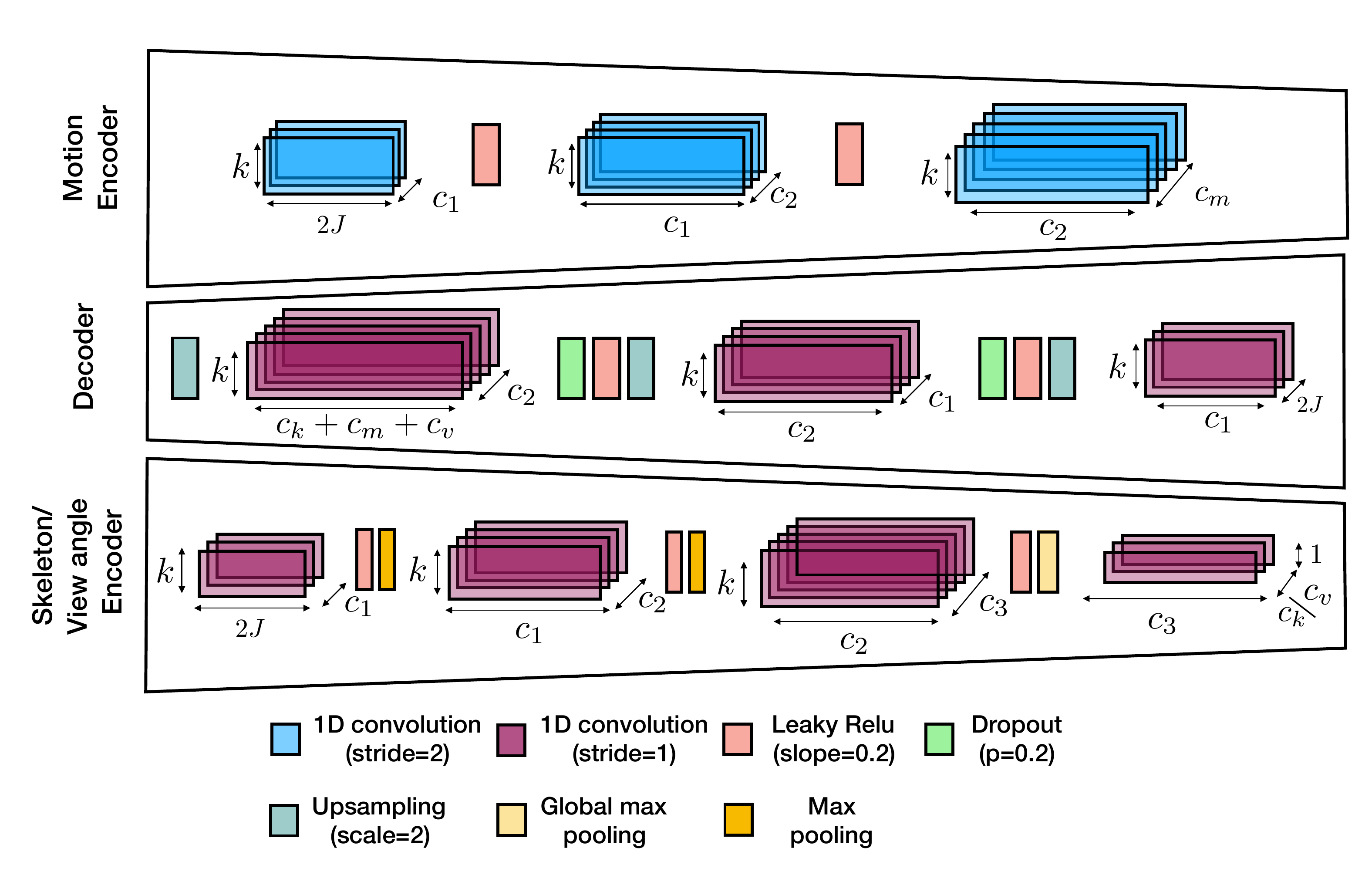} 
	\caption{Our network architecture consists of 3 encoders: motion (top), skeleton and view angle (bottom) and a single decoder (middle). The kernel sizes of the convolution layers are indicated in the figure, and the type of the layers is indicated in the legend at the bottom.
	} 
	\label{fig:layers}
\end{figure}

In practice, our implementation consists of three encoders (motion, skeleton, and view) and one decoder, with the two static encoders (skeleton and view) sharing the same structure. The layers and the dimensions of the different components are shown in Figure~\ref{fig:layers}.

All of our components are based on 1D convolution layers that learn to extract time invariant features from the input sequence (Holden \etal~\shortcite{holden2015learning}), where each layer contains $c_{\text{out}_i}$ kernels of size $k\times c_{\text{in}_i}$. For a detailed description of the parameters of each layer, please refer to the appendix.

In our implementation, the convolution layers in the encoders downsample the temporal axis using stride~2, while in the decoding part we use nearest-neighbor upsampling followed by convolution with stride~1 to restore the temporal information. The reason for the difference is that we found that a symmetric implementation leads to small temporal jittering in reconstruction, a phenomenon that also exists in image generation when a chess board pattern appears in the decoded image \cite{odena2016deconvolution}. In addition, we use Leaky Rectified Linear Units (Leaky ReLU) with slope~0.2, dropout layers ($p=0.2$) to suppress overfitting, and convolution with kernel size~1 to further reduce the number of channels of the fixed size, time independent, latent vector in the static encoders into a smaller one.  In order to optimize the weights of the neural network, based on the loss term in \eqref{eq:total_loss}, we use the AmsGrad algorithm \cite{reddi2018convergence}, a variation of the Adam \cite{kingma2014adam} adaptive gradient descent algorithm.

\section{Supporting Videos in the Wild}

Since our network is trained on clean synthetic data, we next describe how we enhance the training to make the model robust to videos in-the-wild, at test time. This is achieved using augmentation, artificial noise, and data from real videos. The augmentation is applied both for the input and reconstructed output.

\paragraph{Augmentation}
To enrich the observed samples, we apply data augmentation in different ways: 
\begin{enumerate}
\item Temporal Clipping: our model does not require motion clips to have a fixed length, but having a fixed window size during training can improve speed, as it enables the use of batches. Therefore, in every iteration we randomly select the temporal length from the set $T\in\{64,56,48,40\}$. This operation enhances the independence of the static representation on the temporal length of the input sequence. 

\item Scaling: we use various scales, $s\in(0.5,1.5)$, which are equivalent to using different camera distances under the weak-perspective camera model in \eqref{eq:weak_projection}. Note that for cross reconstruction we apply the same scaling to the output that carries the same skeleton attribute, which means that our skeleton size contains the information about scale (namely, two skeletons with different scales will be mapped to different points in the skeleton latent space).

\item Flipping: we left-right flip the joints to obtain augmented skeletons with $\tilde{p}_j^r=\left( -(p^l_j)_x, (p^l_j)_y\right)$, $\tilde{p}^l=\left( -(p^r_j)_x, (p^r_j)_y\right) $, were $p^r_j$ and $p^l_j$ are the left and right positions of a symmetric joint $j$ (e.g, left and right shoulder). Here we apply the same flip to the output that carries the same motion attribute.
\end{enumerate}

\paragraph{Artificial Noise}
Due to the fact that 2D pose estimation algorithms, when applied on videos in the wild, yield results that might contain noise and missing joints, we artificially add noise to the input and dropping joints by randomly $(p=0.05)$ setting their coordinates to zero, while the ground truth output remains complete. Thus, similarly to denoising autoencoders, this operation trains our decoder to perform as a denoiser that returns smooth, temporal coherent sequences, and to cope better with videos in the wild.

\paragraph{Reconstruction of real videos}
During training, we found it helpful to provide the network with some motions that were extracted from videos in the wild. Specifically, in every epoch, we add to the training a set of samples that were extracted from the UCF101 dataset \cite{soomro2012ucf101}, combined with the Penn Action dataset \cite{zhang2013actemes}. The 2D poses were extracted by the method of Cao~\etal~\shortcite{cao2016realtime}. The sequences were split into temporal windows, preprocessed and augmented in the same way, and served as an additional 2000 input samples to the network. Since there are no labels for real videos, for those inputs we apply only the standard reconstruction loss, $\Loss_{\text{rec}}$.



\section{Results and Evaluation}
\label{sec:results}
In this section we report on some experiments that analyze the performance of various components in our framework and present comparisons to state-of-the-art techniques for motion retargeting.

We implemented our network in PyTorch, and performed a variety of experiments on a PC equipped with an Intel Core i7-6950X/3.0GHz CPU (16 GB RAM), and an NVIDIA GeForce GTX Titan Xp GPU (12 GB). Training our network takes about 4 hours. Our dataset was split into two parts, training and validation, with each character and each motion assigned to one of these parts. In other words, there is no overlap between the training and the validation characters and motions. 

\begin{figure}
	\centering
	\includegraphics[width=\linewidth]{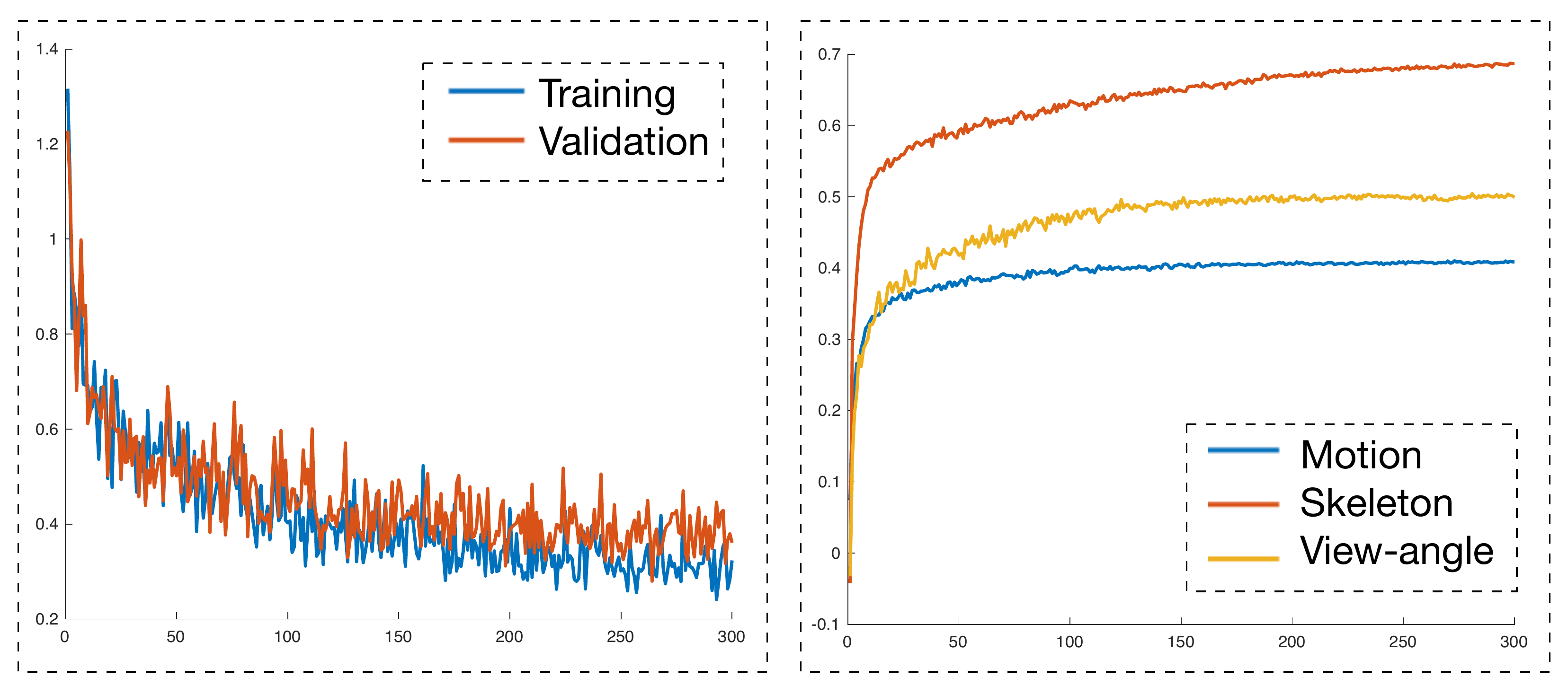} 
	\begin{tabular}{cc}
		(a) \hspace{3cm}&\hspace{3.6cm} (b) 
	\end{tabular}
	\caption{Explicit and implicit learning. (a) Our cross reconstruction loss as a function of the number of epochs for the training (blue) and validation (orange) data. (b) Mean silhouette coefficient of our test set, for the 3 latent spaces (motion, skeleton and view angle). It may be seen that the network learns to cluster the data even though this isn't explicitly required.}
	\label{fig:silhouette_loss}
\end{figure}

\subsection{Ablation study}
\label{sec:ablation}

In order to examine the performance of the cross reconstruction loss, $\Loss_{\text{cross_rec}}$, we first train our network using only this loss. Figure~\ref{fig:silhouette_loss}(a) plots the loss curve as a function of the number of epochs, applied to training (blue) and validation (orange) data, demonstrating that the network generalizes well and does not overfit the training data. Next, we show that with this loss the network implicitly learns to cluster the input, despite the fact that it imposes no explicit requirement for separation in the latent space.
To measure the ability to cluster, we use the mean silhouette coefficient \cite{kaufman2009finding}, given by 
\begin{equation}
\bar{S}_{M}=\frac{1}{|\mm||\ms|}\sum_{i\in\mm}\sum_{j\in\ms}S_M(\bp_{i,j}),
\end{equation}
where
\begin{equation}
S_M(\bp_{i,j}) = \frac{B(E_M(\bp_{i,j})) - A(E_M(\bp_{i,j}))}{\max \left\lbrace A(E_M(\bp_{i,j})),\; B(E_M(\bp_{i,j})) \right\rbrace}.\nonumber
\end{equation}
Here $A(E_M(\bp_{i,j}))$ is the average distance between $E_M(\bp_{i,j})$ and all other samples within the same cluster, 
while $B(E_M(\bp_{i,j}))$ is the smallest average distance between $E_M(\bp_{i,j})$ to all the points in any other cluster. 
The clustering of the skeletons and the view parameters is measured in the same manner, by evaluating $\bar{S}_{C}$ and $\bar{S}_{V}$ for the corresponding latent spaces.

\begin{figure}
\centering
\includegraphics[width=\linewidth]{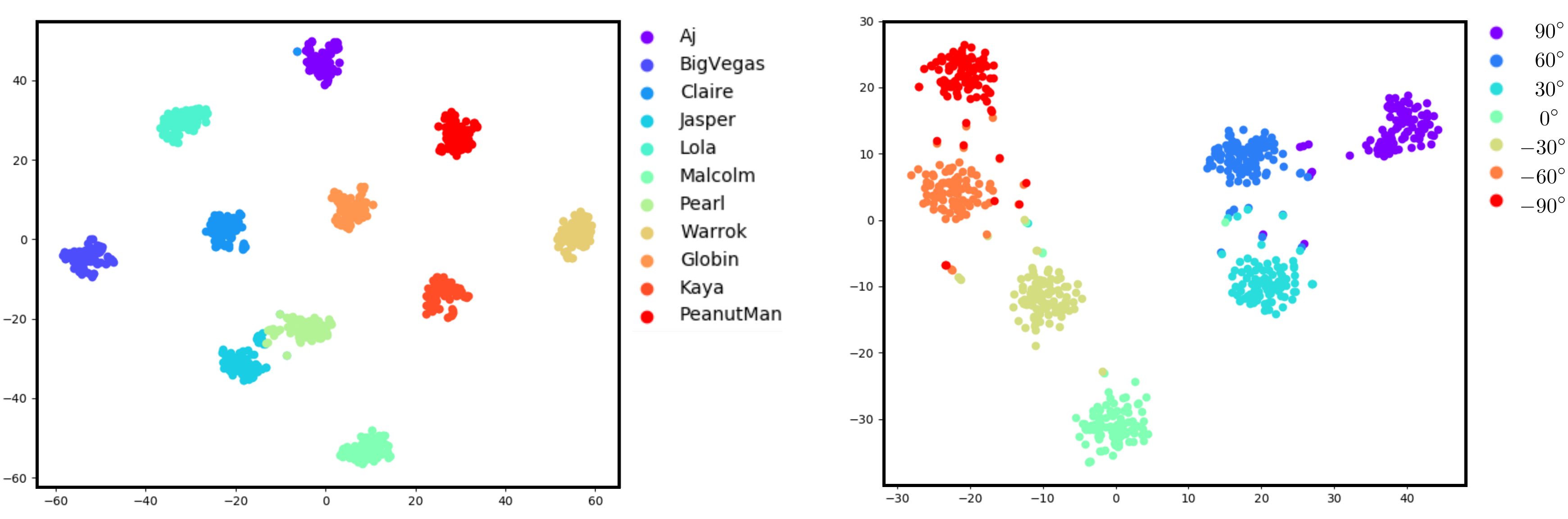} 
\begin{tabular}{cc}
(a) \hspace{4.8cm}&\hspace{3.8cm} (b) 
\end{tabular}
\caption{Latent clusters using the cross reconstruction loss, $\Loss_{\text{cross_rec}}$. The samples of our test set are encoded into the latent spaces, visualized using t-SNE. (a) Skeleton latent space labeled by character name. (b) View latent space labeled by view angle.}
\label{fig:clusters_body_view}
\end{figure}

After each epoch we calculate the mean silhouette coefficient of the latent representation of our test set (derived from the validation set and containing 11 characters, 15 motions and 7 view-angles.  Figure~\ref{fig:silhouette_loss}(b) plots the mean silhouette coefficient for each of the three latent spaces as a function of the number of epochs. The coefficients are increasing, which indicates that the network implicitly learns to cluster the labeled groups, even though this isn't explicitly required. The resulting latent spaces of the skeleton and view angle (after 300 epochs) are shown in Figure~\ref{fig:clusters_body_view}, visualized using t-SNE \cite{maaten2008visualizing}. It may be seen that the samples are well clustered, in both latent spaces. The samples in the view latent space are more scattered, since the view angle is expressed in a skeleton-centric coordinate system, which is averaged over the poses of a given motion (Section \ref{subsec:dataset}). Thus, there's a dependency between the motion and the coordinate system, which gives rise to a larger variance in the view-dependent static latent parameters among different sequences that share the same view-angle label.

\begin{figure}
\centering
\includegraphics[width=\linewidth]{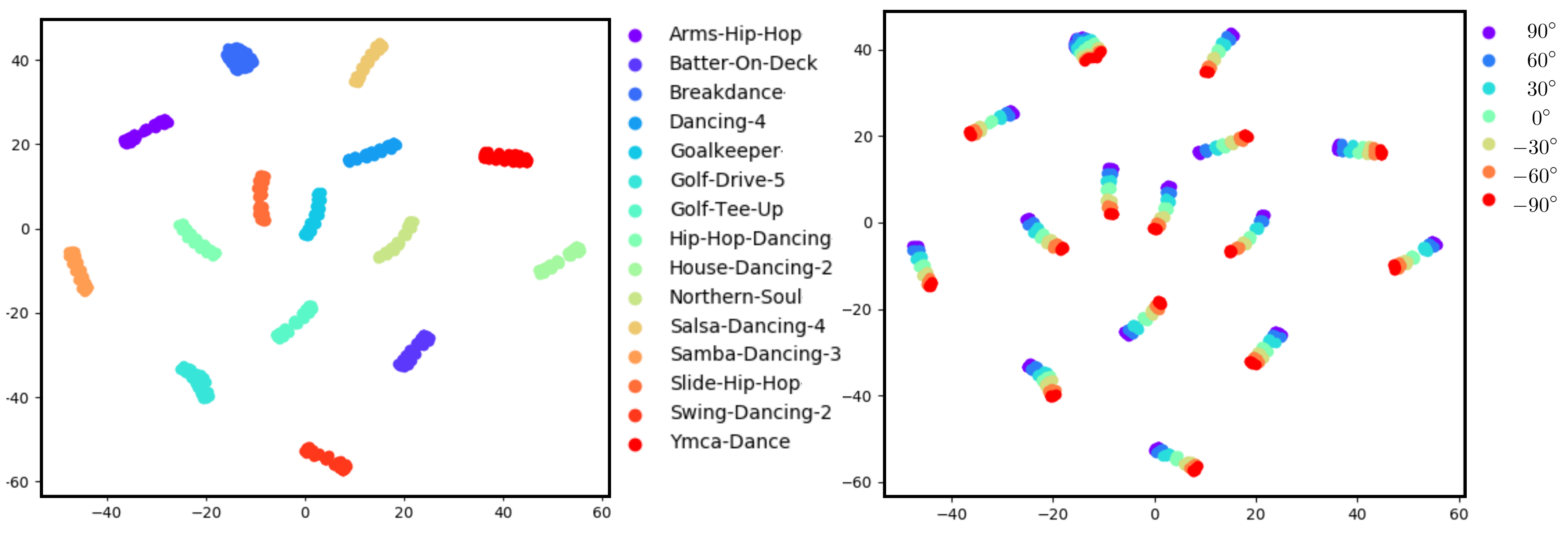} \\
\begin{tabular}{cc}
(a) \hspace{4.8cm}&\hspace{3.8cm} (b) 
\end{tabular}
\caption{Motion latent space clusters using the cross reconstruction loss, $\Loss_{\text{cross_rec}}$. Samples of the test set are encoded in the the motion latent spaces and demonstrated in 2D (using t-SNE). (a) Motion latent codes labeled by motion (b)
Motion latent codes labeled by view angle. It can be seen that the network learns well how to cluster different motions, while each cluster contains a view angle-dependent structure.}
\label{fig:clusters_motion}
\end{figure}

The motion latent space for the same setup (training using only $\Loss_{\text{cross_rec}}$) is visualized in Figure~\ref{fig:clusters_motion}(a). It may be seen that the different motions also become clearly clustered. Interestingly, when labeling each sample using its view-angle label in Figure~\ref{fig:clusters_motion}(b), a clear inter-cluster structure emerges, revealing that the motion latent space encodes some information about the view angle as well. 
This may also be attributed to the dependency mentioned above.
Thus, the large variation between different view angle projections can't be totally disentangled from the motion, using only our cross reconstruction loss.

\begin{figure}
\centering
\includegraphics[width=\linewidth]{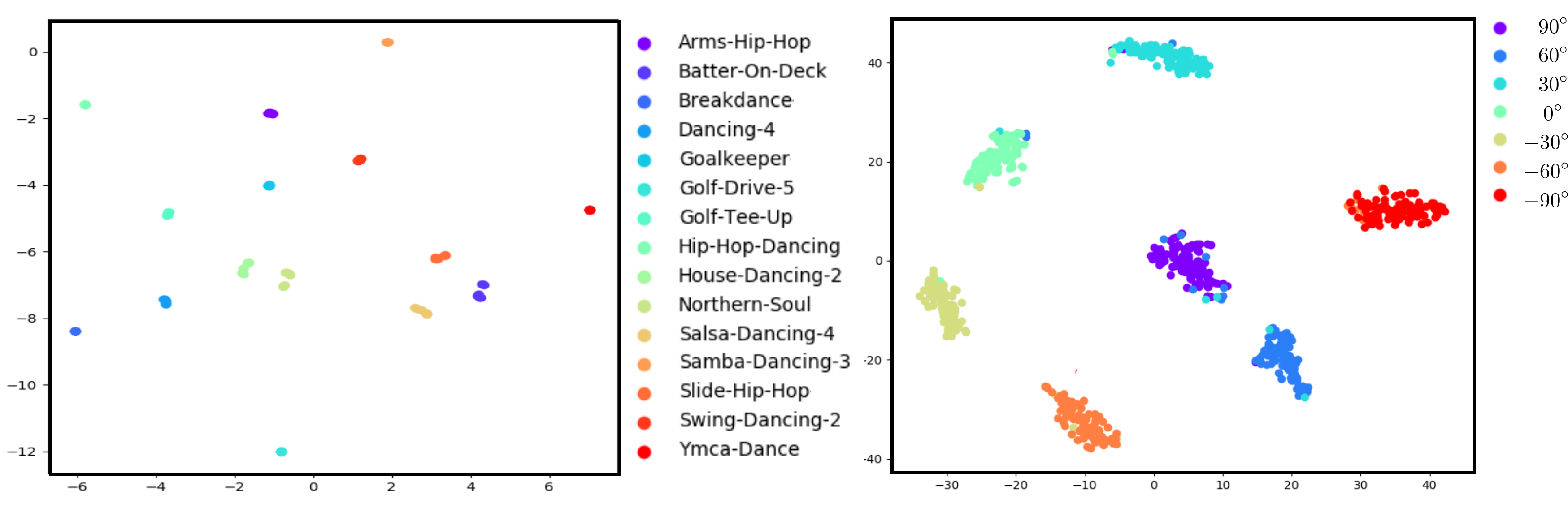} 
\begin{tabular}{cc}
(a) \hspace{4.8cm}&\hspace{3.8cm} (b) 
\end{tabular}
\caption{Latent clusters using the cross reconstruction loss and the triplet loss, $\Loss_{\text{cross_rec}}+\Loss_{\text{trip}}$. Samples of the test set are encoded in the the latent spaces, and visualized in 2D (using t-SNE). (a) Motion latent space labeled by motion (b) View latent space labeled by view angle.}
\label{fig:clusters_triplet_motion_view}
\end{figure}

As explained in Section~\ref{subsec:decompose}, in order to impose disentanglement between the attributes, we make use of the triplet loss. Figure~\ref{fig:clusters_triplet_motion_view} demonstrates the contribution of $\Loss_{\text{trip}}$ to the clusters of the motion and the view angle. It can be seen that the clusters become tighter, which is also echoed by higher silhouette scores after training ($\bar{S}_M=0.45$, and $\bar{S}_C=0.75$, $S_V=55$ with triplet loss, versus $\bar{S}_M=0.39$, and $\bar{S}_C=0.69$, $S_V=48$ without).

The top part of Table~\ref{tab:retargeting} reports the MSE error (defined in the next section in Eq.~\eqref{eq:mse}) between the retargeted output of pairs from the validation set and the ground truth. The results show that the inclusion of the triplet loss, which enhances the disentanglement of the three attributes, also improves the retargeting performance. On the other hand, using only an ordinary reconstruction loss with the triplet loss $\Loss_{\text{rec}} + \Loss_{\text{trip}}$ significantly degrades the re-composition performance, and cannot properly perform the retargeting task.

It can be concluded that our cross reconstruction loss $\Loss_{\text{cross_rec}}$, which implicitly trains the network to efficiently cluster the data in each of the latent spaces, is the the most crucial term for the retargeting task. The triplet loss further enhances the tightness of the clusters and imposes better disentanglement of the latent features, which further improves the retargeting performance.




%
%


\begin{table}
	\caption{Quantitative comparisons. The top portion of the table reports the MSE that our framework achieves on our test dataset, under different loss terms. The bottom portion reports the MSE scores of other retargeting methods, on the same dataset.}
	\label{tab:retargeting}
	\begin{minipage}{\columnwidth}
		\begin{center}
			\begin{tabular}{ll}
				\toprule
				Method & MSE\\
				\midrule
				Ours: $\Loss_{\text{cross_rec}}$ + $\Loss_{\text{triplet}}$ & {\bf 1.23} \\
				Ours: $\Loss_{\text{cross_rec}}$ & 1.44 \\
				Ours: $\Loss_{\text{rec}}$ + $\Loss_{\text{triplet}}$  & 11.97 \\
				\midrule
				Naive 2D forward kinematics & 3.44 \\
				NKN \cite{villegas2018neural} & 1.91 \\
				3D baseline (naive) & 2.25 \\
				3D baseline (rescaled velocity) & 0.2 \\
				\bottomrule
			\end{tabular}
		\end{center}
	\end{minipage}
\end{table}

\begin{figure*}
	\centering
	\begin{tabular}{ccccc}
		\hspace{0.5cm} Motion-Input & \hspace{1.7cm}  Static-Input & \hspace{2.0cm} 2D-FK  &  \hspace{1.1cm} NKN \cite{villegas2018neural}  & \hspace{1.2cm} Ours \hspace{3.5cm}
	\end{tabular}
	\includegraphics[width=\linewidth]{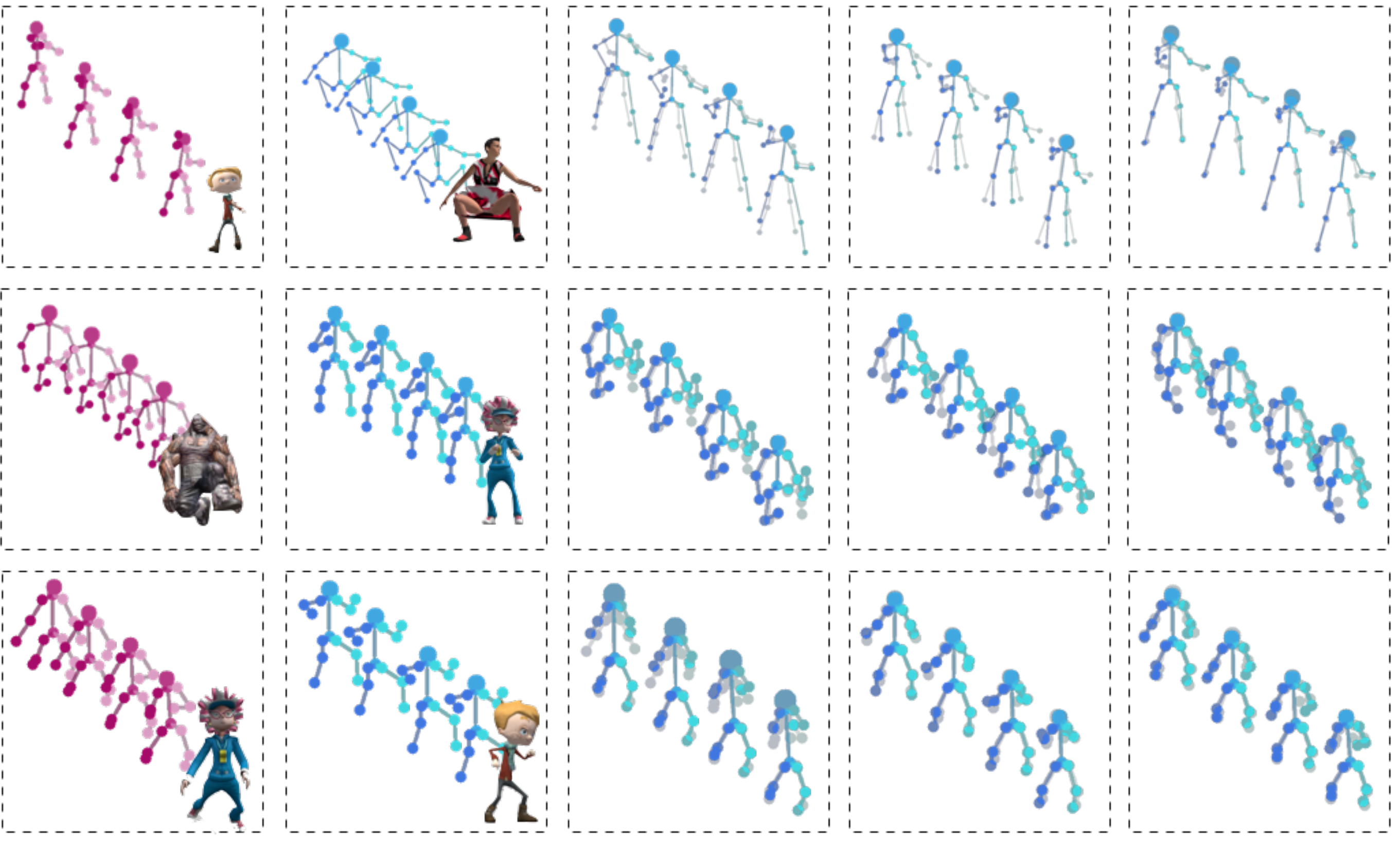} 
	\caption{Comparison to other retargeting methods. Given a motion input sequence (first column) and a sequence from which static parameters are extracted (second column), the results of three retargeting approaches are shown in the right columns: 2D Forward Kinematics (2D-FK, middle column), Neural Kinematic Networks (NKN) \cite{villegas2018neural} (4th column) and our method (rightmost column). The ground truth is depicted in light gray on top of every output.
	}
	\label{fig:comparison_character}
\end{figure*}

\subsection{Comparison}

In this section, we report two experiments for evaluating our method against other motion retargeting algorithms. First, we compare several methods under a scenario where the ground truth 3D poses are available. This is done using the synthetic animated 3D characters from our validation set. Second, we compare the methods under the more realistic scenario, where the motion to be retargeted is captured by a video, without the benefit of exact 3D poses. The latter scenario is the one targeted by our approach.

While there is a variety of optimization-based approaches that perform motion retargeting by solving an inverse-kinematics problem, most of these methods expect the user to provide motion specific constraints or goals, which is not feasible to be done on a large scale. Thus, our method is compared with the state-of-the-art method of Villegas \etal~\shortcite{villegas2018neural}, which performs unsupervised 3D motion retargeting via neural kinematic networks (NKN) and was also trained on synthetic (3D) motion data obtained from Mixamo \cite{mixamo}. 

In addition, we compare to a naive approach that applies 2D retargeting directly on the 2D input (2D Forward Kinematics), resembling a naive 3D retargeting, where the length of the limbs is modified to match the target skeleton, while preserving the joint angles. In 2D, a per-limb scaling is applied to the source character so the average of each limb length over time is equal to the average length of the corresponding limb of the target character. The rescaled limb length at time $t$ of the resulting motion is given by 
\begin{equation}
\hat{l}_{\text{out},j}^{(t)}= \frac{T_{\text{src}}\sum_{i=1}^{T_{\text{tgt}}}l_{\text{tgt},j}^{(i)}}{T_{\text{tgt}}\sum_{i=1}^{T_{\text{src}}} l_{\text{src},j}^{(i)}}l_{\text{src},j}^{(t)},
\end{equation}
where $l_{\text{src},j}^{(t)}$ and $l_{\text{tgt},j}^{(t)}$ are the lengths of $j$th limbs in time $t$ of the source and target characters, and $T_{\text{src}}$ and $T_{\text{tgt}}$ are the temporal lengths of the source and target sequence, respectively. The joint positions are then calculated, based on the limb lengths, from the root to the end-effectors (in a tree structure), while preserving the 2D  angles, between connected limbs, of the original pose. The global velocity is rescaled based on the ratio between the average heights of the skeletons.

Finally, we include two 3D baselines for retargeting. The naive variant of this baseline directly copies the per-joint rotations (quaternions), as well as the global velocity from the input motion to form the retargeted motion. A more sophisticated variant also rescales the global velocities based on the ratio between skeleton heights.

\paragraph{Retargeting with exact 2D and 3D poses}
All the results in this experiment are compared against the ground truth, which is in practice based on the 3D motion retargeting algoritm Human-IK of AutoDesk \cite{montgomery2012tradigital}.
Since motion retargeting methods are performed in the 3D domain, while our method operates directly on the 2D joint positions,
we project the ground truth, as well as the results from the 3D methods (Villegas \etal~\shortcite{villegas2018neural}, 3D baseline) into 2D using the same camera parameters that were used to project the corresponding motions in our dataset.

The error between the output and the ground truth is calculated as the MSE between corresponding joint positions over time. Since large characters tend to produce larger deviations, we normalize the error by the character's 3D height (calculated by summing the lengths of the leg, torso and neck). The error term is given by
\begin{equation}
E \left( \bp_{i,j}, \hat{\bp}_{i,j} \right)  = \frac{1}{h_j}\frac{\Vert \bp_{i,j} - \hat{\bp}_{i,j}\Vert^2 }{2JT},
\label{eq:mse}
\end{equation}
where $h_j$ is the height of the character $j\in \mathcal{C}$.
In this experiment we used pairs of 3D characters from our test dataset. Table~\ref{tab:retargeting} (bottom part) reports the resulting errors, and a few visual examples are shown in Figure~\ref{fig:comparison_character}. It may be seen that our method yields better results than the naive 2D forward kinematics approach, which scales the limbs based on the average length, resulting in erroneous joint positions, especially in sequences with large changes in the projected length of individual limbs. In addition, it may be seen that the error of Villegas \etal~\shortcite{villegas2018neural} is larger than our method's, but it should be noted that their method is unsupervised, while ours is.

Finally, an analysis of the error of the 3D baselines, reveals that most of the error in the naive version is attributed to the global motion part. Computing the local error (by subtracting the root position) yields a much smaller error of 0.09. This makes sense, since the ground truth is a result of an optimization algorithm which first rescales the limbs, and then optimizes the joint positions by imposing physical constraints, which have a significant effect on the global position (especially the foot contact constraint). In comparison, using the 3D baseline with velocity rescaling, yields an error of 0.2, achieving higher accuracy than our method.
Thus, we conclude that, run-time considerations aside, given the full 3D representation of the source motion and the target character, classic IK-optimization methods are able to perform better on the task of 3D motion retargeting.
However, we next show that the situation is different when the motion is captured by a video.

\paragraph{Retargeting of video-captured motion}
In our second experiment, we perform a quantitative comparison using synthetic videos of characters from Mixamo \cite{mixamo}, and a qualitative comparison using videos in the wild, for which no ground truth is available. These videos include a subset from the UCF101 dataset \cite{soomro2012ucf101}, as well as several videos from YouTube.

The comparison is done against the full 3D pipeline that is outlined in Figure~\ref{fig:bypass_3d}(b). Given a pair of videos, we first apply a 3D pose estimation method, then perform the retargeting in 3D, and finally project the motion back to 2D using the estimated camera parameters.
The 3D retargeting was done with the 3D baseline (with velocity rescaling), as it was able to achieve the best results in the previous experiment. 

We used two state-of-the-art algorithms for 3D pose estimation, that are suitable for videos. The first method is VNECT~\cite{mehta2017vnect}, which recovers a full, global, 3D skeletal pose of a human per frame, and then uses inverse kinematics to fit a single skeleton to the recovered joint positions in a temporally consistent manner. In this comparison we used the official code supplied by the authors, which doesn't contain the temporal fitting part, and smoothed the resulting joint positions with a Gaussian kernel. The second is HMR~\cite{kanazawa2018end}, extended to videos by applying a temporal coherence optimization of Peng \etal~\shortcite{peng2018sfv}. HMR was also used to estimate the camera parameters in both 3D pipelines.

\begin{figure*}
\centering
\begin{tabular}{cccc}
\hspace{-0.2cm} Inputs & \hspace{2.1cm} HMR \cite{kanazawa2018end} &  \hspace{2.0cm} VNECT \cite{mehta2017vnect} & \hspace{1.5cm} Ours
\end{tabular}
\begin{tabular}{cccc}
 3D-Reconstruction \hspace{0.4cm} & 3D-Retargeting \hspace{1.0cm}  & \hspace{-0.4cm} 3D-Reconstruction &  \hspace{0.3cm} 3D-Retargeting 
\end{tabular}
\includegraphics[width=\linewidth]{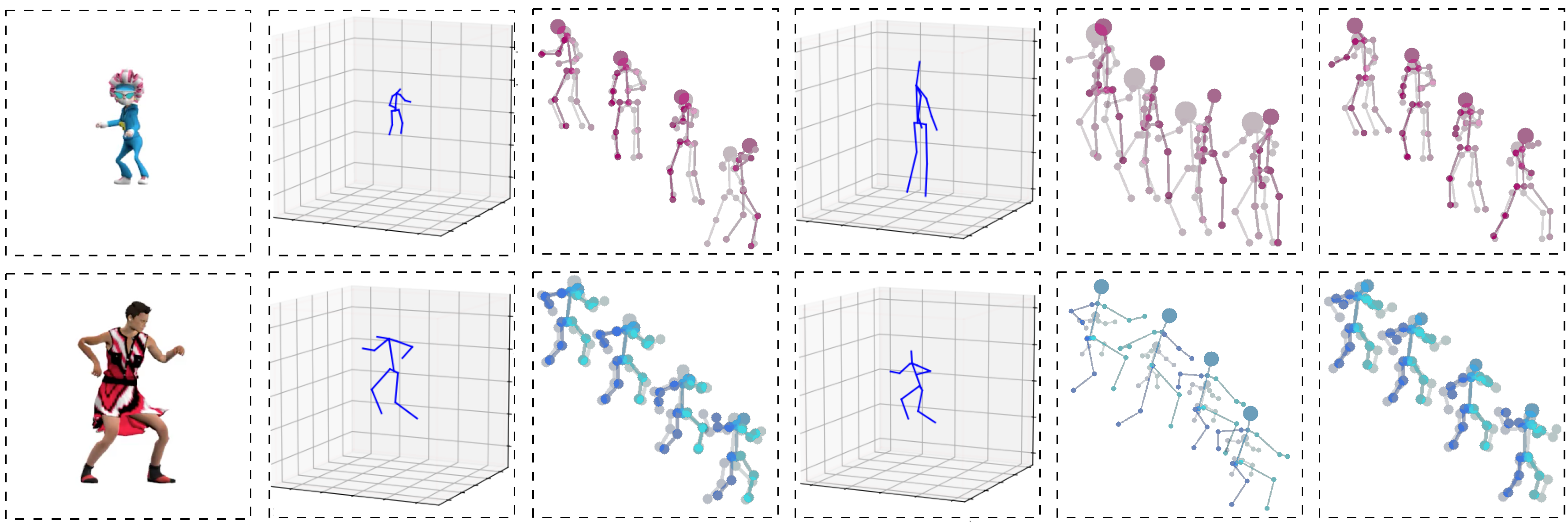}\\
(a)\\ 
\begin{tabular}{cccc}
\hspace{-0.2cm} Inputs & \hspace{2.1cm} HMR \cite{kanazawa2018end} &  \hspace{2.0cm} VNECT \cite{mehta2017vnect} & \hspace{1.5cm} Ours
\end{tabular}
\begin{tabular}{cccc}
 3D-Reconstruction \hspace{0.4cm} & 3D-Retargeting \hspace{1.0cm}  & \hspace{-0.4cm} 3D-Reconstruction &  \hspace{0.3cm} 3D-Retargeting 
\end{tabular}
\includegraphics[width=\linewidth]{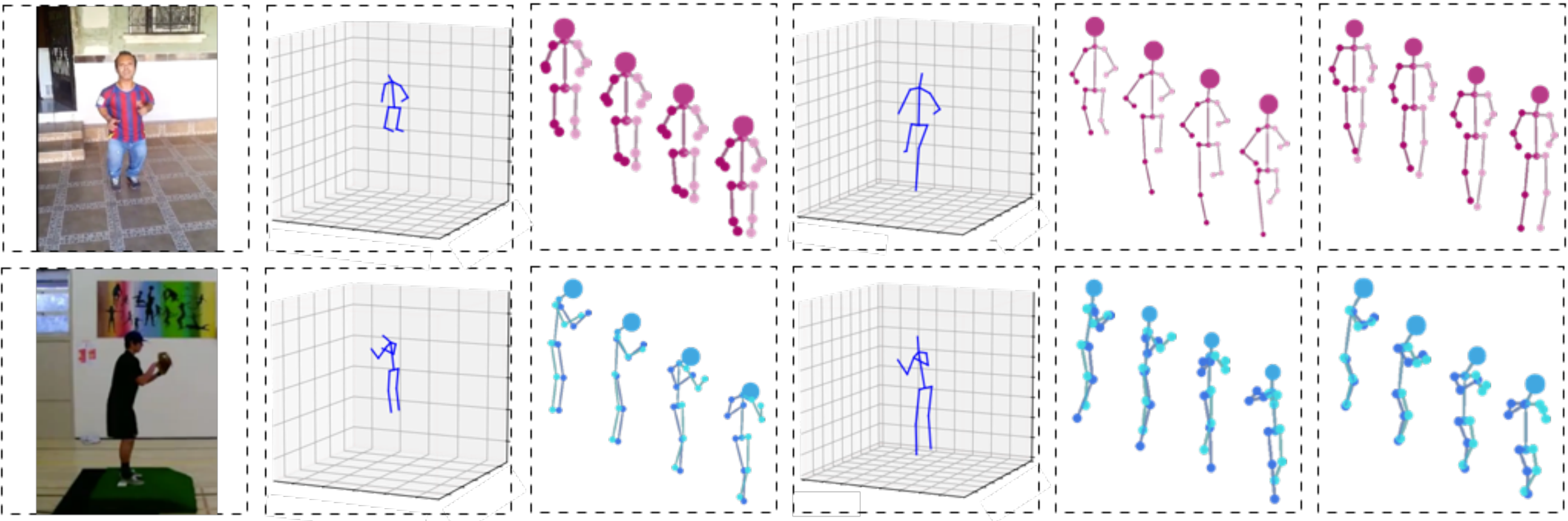}\\
(b)\\ 
\caption{Comparison to retargeting of video motion in 3D (using 3D pose estimation). From left to right: Original input videos, reconstructed 3D pose and retargeted poses using HMR, reconstructed 3D pose and retargeted poses using VNECT, our results, retargeted directly in 2D.}
\label{fig:qualitative_3d}
\end{figure*}

The retargeting results may be found in the supplementary video.
Selected frames from these results are shown in Figure~\ref{fig:qualitative_3d}. It may be seen that for some of the examples VNECT yields temporally inconsistent joint positions which result in unnatural motions, and consequently struggles to accomplish the retargeting task.
Another root problem, in most cases, is the wrong scale of the skeleton. There is an ambiguity between the skeleton and the camera, since multiple combinations of skeleton size and camera parameters may yield the same projection. Unfortunately, VNECT does not recover its own camera parameters, and we use HMR to recover them. The wrong scale interferes with the retargeting, but the ambiguity causing this issue cannot be resolved, unless we know the ratio between the heights of the characters, which is unknown for videos in the wild.  

As for HMR, despite the fact that it recovers a skeleton whose 2D projection is correct, it may be seen that the 3D joint positions might be incorrect, especially for characters with unusual limb proportions.
Figure~\ref{fig:qualitative_3d} demonstrates such an example, where the reconstructed legs of a person of small stature in the video are unnaturally bent. This leads to a retargeted result where the skeleton of the target individual also has bent legs, which appear unnatural in the 2D projection. In contrast, our method generates a 2D projection where the bottom part of the target individual's leg appears to have normal length.

Since for the Mixamo videos, ground truth retargeted motions are available, we are able to report quantitative results for these videos. The HMR method achieves an error of 2.08, while our method achieves a lower error of 1.70.

\section{Applications}
\label{sec:results}
Having the ability to extract and retarget human motion directly from videos paves the way to a variety of applications.

\begin{figure*}
\centering
\begin{tabular}{cccc}
\hspace{-0.0cm} Reference Video & \hspace{2.5cm}  Driving Video & \hspace{2.2cm} Global Scaling  &  \hspace{2.3cm} Our Retargeting
\end{tabular}
\includegraphics[width=\linewidth]{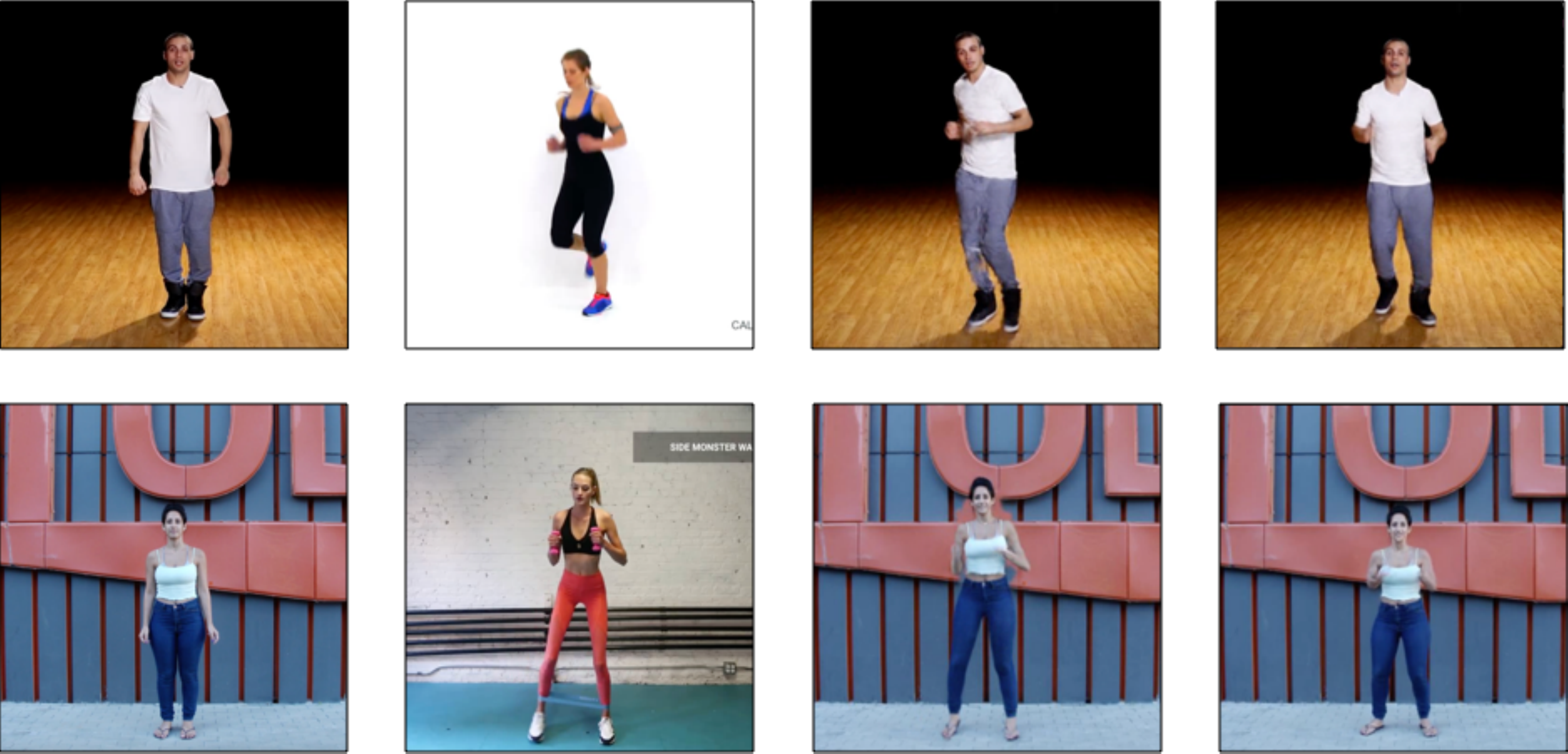} 
\caption{Using the method of Chan \etal~\shortcite{chan2018everybody}, we train a network to regenerate frames depicting an actor in a reference video (left) based on 2D poses extracted from a driving video (second column). A simple global scaling leads to erroneous proportions and artifacts (third column), while a generation using our retargeting method yields correct body proportions and adaptation of the original orientation (right), leading to plausible results.}
\label{fig:performance_cloning_1}
\end{figure*}

\subsection{Performance Cloning}
The ability to perform motion retargeting in 2D enables one to use a video-captured performance to drive a novel 2D skeleton, with possibly different proportions. This is analogous to the 3D domain, where an articulated 3D character, which is already rigged for animation, can be animated by retargeting of captured or animated driving performance.

Recently, several performance cloning techniques proposed deep generative networks, trained to produce frames that contain the appearance of a target actor reenacting the motion of a driving actor \cite{chan2018everybody,aberman2018deep,liu2018neural}.

While Liu \etal~\shortcite{liu2018neural} require a 3D mesh as a prior to the network, Chan \etal~\shortcite{chan2018everybody} and Aberman~\etal~\shortcite{aberman2018deep} use a 2D skeleton as a prior. In order to retarget the skeleton of the driving actor to fit the dimensions of the target actor, both methods use global scaling and translation. This approach limits the system to work with actors that share the same skeleton proportions and that were captured from similar view angles.
In order to demonstrate the benefit of our method for that task we use the technique of Chan et al.~\shortcite{chan2018everybody}, and train a network on a given reference video, which learns to generate frames from 2D poses. However, instead of using the global scaling, we generate the sequence of 2D poses using our method, by recomposing the motion extracted from a video of the driving actor with the skeleton and the view angle extracted from a reference video of the target individual.

For example, we trained the aforementioned framework on a 3-minute video from YouTube\footnote{https://www.youtube.com/watch?v=nzta5cy2jE0}.
This video depicts a frontally-captured male dancer demonstrating hip-hop moves. After training, the model is driven by another video depicting a female fitness trainer, with different proportions, who is not frontally captured. The top row of Figure~\ref{fig:performance_cloning_1} shows frames from the reference video, driving video, global scaling result, and our result. Using only global scaling, it is impossible to properly generate frames of the dancer performing the motion from the driving video. It may seen that the proportions between the male dancer's upper part and the lower part were modified to match those of the female trainer. Furthermore, the generated frames contain various artifacts, since the network didn't see the dancer in this orientation during training.  However, with our 2D retargeting technique, the body proportions are properly rescaled, and the frames may be rendered from a frontal view, yielding a plausible video of the dancer reenacting the motions in the driving video.
The bottom row of Figure~\ref{fig:performance_cloning_1} shows another example, where the body proportions of the two characters are very different.

\subsection{Motion Retrieval}

Using our motion representation, we can search in a dataset of videos in-the-wild for motions similar to one in a video given as a query, with the search being agnostic to the body proportions of the individual and the camera view angle.  
Furthermore, since our latent motion representation contains a temporal axis that preserves the temporal information (up to the receptive field of the network), the searched videos may have different temporal lengths, with the results localizing the (shorter) query motion inside the retrieved sequences.

We demonstrate a motion search engine that enables to efficiently search for a query motion in a dataset of videos. When adding a video into the dataset, the system passes it through the 2D pose estimation component \cite{cao2016realtime}, then extracts its latent motion representation by a forward pass through the trained motion encoder, $E_M$. The resulting latent motion representations of the different videos are concatenated along the temporal axis, and are saved in the dataset in this form.

Given a video containing a query motion, we extract the motion representation as described above, and search for the maximal cross-correlation between the query and the concatenation of the motions in the dataset, using a single convolution pass. Once the best match has been found, the corresponding piece of video is trimmed and returned. Since the search is performed on the latent representation, the engine enables to localize the retrieved motions with a temporal accuracy up to the receptive field ($r=12$). The performance of the search is of $O(N)$, where $N$ is the number of videos in the dataset, but can be improved with more efficient search strategies. 

In our experiments, we applied the search over a set of videos in-the-wild from the UFC101 dataset \cite{soomro2012ucf101}, combined with the Penn Action dataset \cite{zhang2013actemes}. The query motions were taken out of these datasets, and depict various actions that are not necessarily contained among those in the datasets.

\begin{figure*}
\centering
\includegraphics[width=\linewidth]{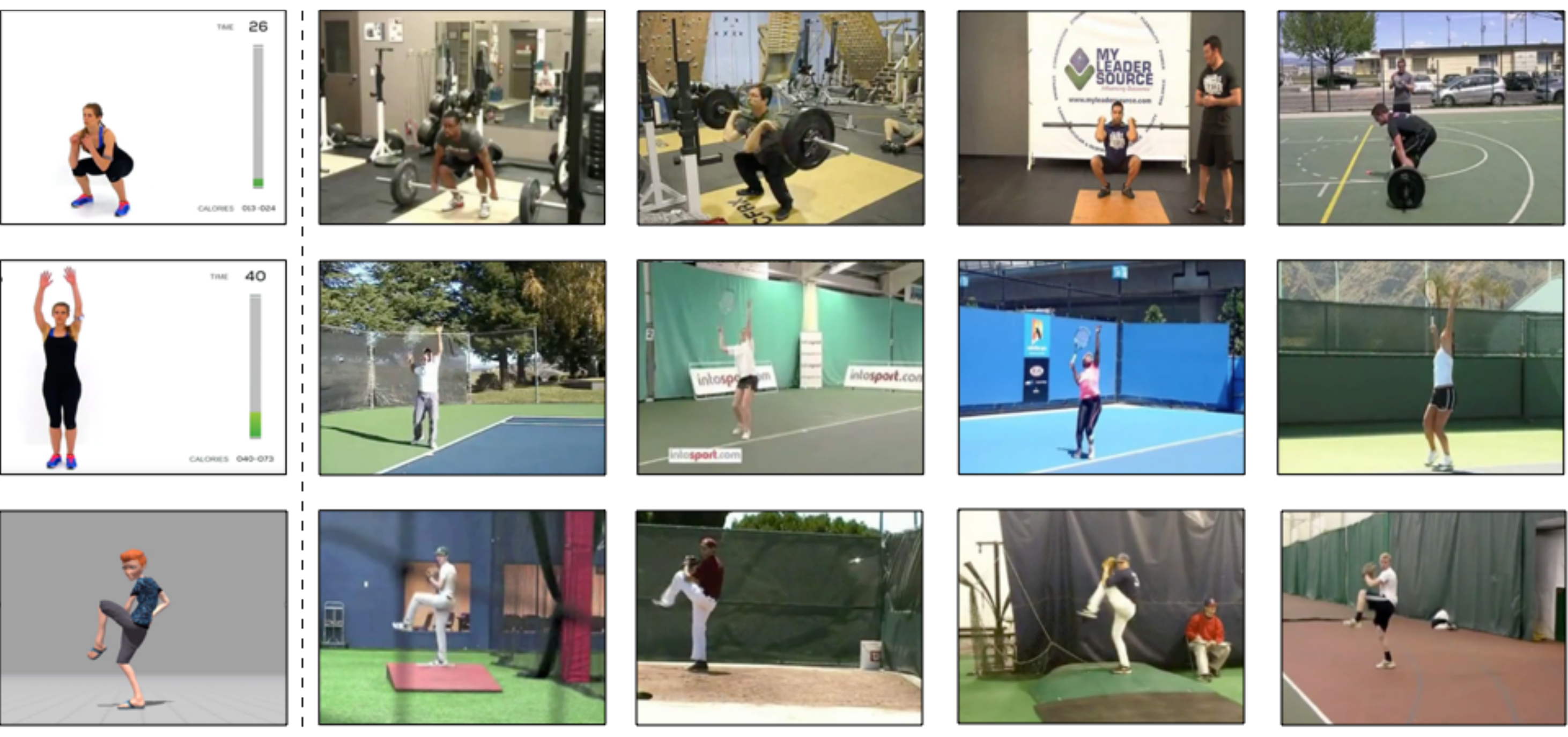} 
\caption{Video-based motion retrieval. The left column shows a single frame from a short video query depicting a motion. The other four columns show frames from the top four results retrieved by our system.}
\label{fig:retrival}
\end{figure*}

Figure~\ref{fig:retrival} shows several examples of short query sequences (left column) and the top four results retrieved by our search (the four other columns). It may be seen that our method is able to find videos that exhibit similar motions to the one in the query, and temporally localize them inside sequences in the database. Note that the retrieved results exhibit a variety of body shapes and view angles, demonstrating the agnosticism of our method to these attributes. In addition, even when the query exhibits a motion that is not identical to those in our dataset, the retrieved motions feature similar limb gestures. For example, a motion where both arms are raised retrieves videos of a tennis serve.

%

\section{Discussion and Future work}
\label{sec:discussion}

We have presented a technique for analyzing video-captured motion, which enables to perform motion retargeting directly on the 2D projections of skeletons, bypassing the notorious problem of lifting the data to 3D. Our framework uses a deep network which is trained on synthetic data to learn to separate the observed motion a dynamic part (the motion) and static parts (the skeleton and the view angle). Our results show that deep networks can constitute a better solution for sub-tasks, such as 2D retargeting, which do not necessarily require a full 3D reconstruction.

Interestingly, loosely speaking, the latent motion remains an elusive intangible representation. It does not possess a meaningful visual representation, unless applied to a specific skeleton. Nevertheless, the motion representation is flexible in the sense that it can represent motion of any duration. 
Moreover, the motion which we refer to as ``character-agnostic'' is in fact also ``view-agnostic'', and it can be combined with an arbitrary skeleton and projected into 2D from arbitrary view directions, assuming it belongs to the  set of views that the network was trained with.

\begin{figure}
	\includegraphics[width=\linewidth]{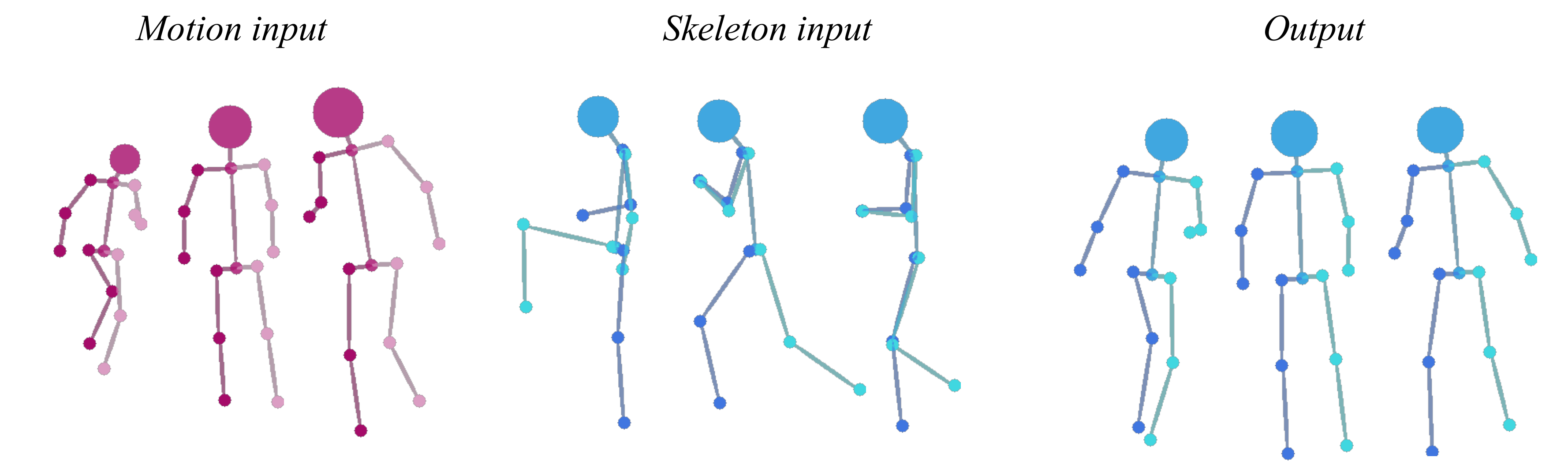} 
	 \\ (a)
	\includegraphics[width=\linewidth]{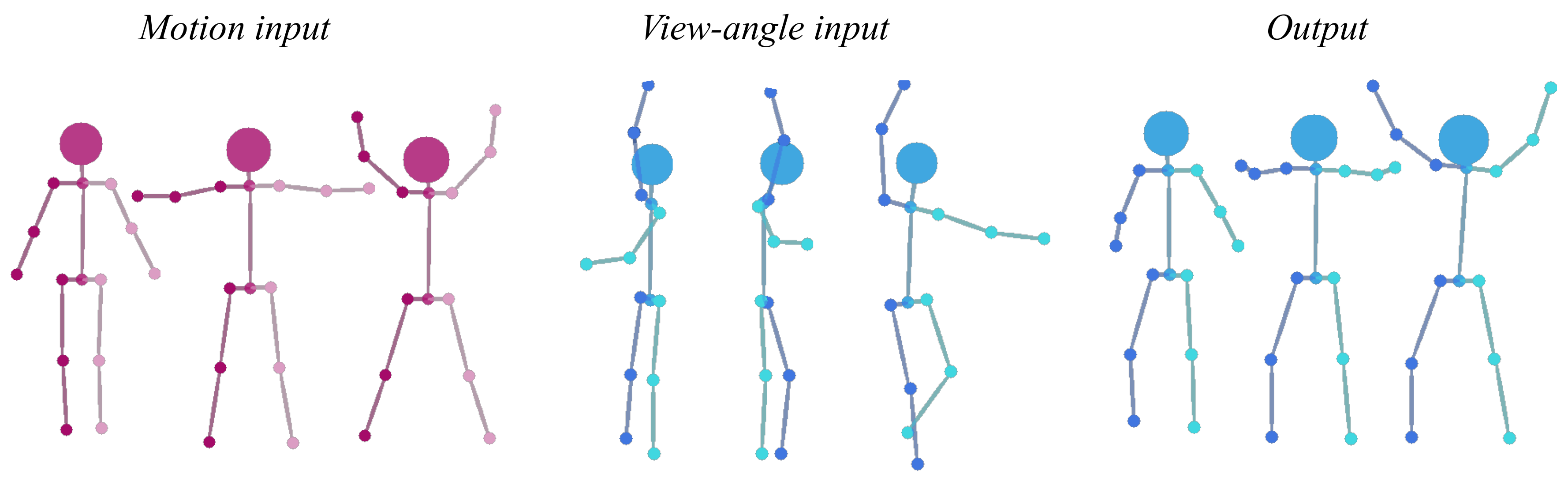} 
	 \\ (b)
	\caption{Failure cases. Our method fails to transfer large scale or view angle variation to the retargeted output motion. (a) Large scale variation. (b) Large view angle variation.}
	\label{fig:failure}
\end{figure}

As a byproduct of our training, latent spaces are generated, where the latent codes tend to cluster. We have shown that applying clustering losses to tightening the clusters can further improve the results. This opens more interesting questions as whether we can have more control on learning these clusters to create better disentangling of the motion data. On the other hand, tight clustering in latent space is not always a virtue. Non tight clusters allow some natural flexibility that may capture better some drifts in the data. For example, currently we assume that the video is captured from a static camera modeled by a weak-perspective transformation. As a result, decomposing long motions that exhibit large variation in the 2D scale or in the view angle may result in artifacts during reconstruction, as can be seen in Figure~\ref{fig:failure} where sequence A (top row) fails to transfer its motion to the retargeted output and sequence B fails to transfer the view angle (bottom row). In the future, we would like to allow larger camera motion, and controlling the camera view latent space is one approach that we are considering. 

Another intriguing problem for future work is to consider using this motion analysis to assist the reconstruction of 3D skeleton from video. In this work, we argued the advantages of bypassing the need to go 3D, but at the same time, being view-agnostic implicitly implies that the 3D data in latent in the network. This provides the motivation to look for means of consolidating the 3D information into a 3D representation. The hope is that it can, at least, improve current methods that estimate 3D poses from video.

\begin{acks}
We thank Andreas Aristidou for his valuable input and the anonymous reviewers for their constructive comments. This work was supported by China National 973 Program (2015CB352501). In addition, partial  support was provided by the Israel Science Foundation (2366/16) and the ISF-NSFC Joint Research Program (2217/15, 2472/17).
\end{acks}

\bibliographystyle{ACM-Reference-Format}
\bibliography{references} 

\appendix
\section{Network Architecture}
The full architecture of our network is summarized in the table below, where \texttt{Conv, LReLU, MP, AP, UpS} and  \texttt{DO} denote convolution, leaky ReLU, max pooling, average pooling and upsampling layers, respectively. All of the convolution layers use reflected padding. $k$ is the kernel width, $s$ is the stride, and the number of input and output channels is reported in the rightmost column.

\begin{center}
  \begin{tabular}{| l | l | c | c | c |}
  \hline
  \toprule
      Name & Layers & $k$ & $s$ & in/out \\
  \hline
  \midrule 
      Motion  & \texttt{Conv + LReLU}  & $8$ & 2 & $30/64$\\
      Encoder & \texttt{Conv + LReLU} & $8$ & 2 & $64/96$\\
      & \texttt{Conv + LReLU} & $8$ & 2 & $96/128$\\
  \hline
  \midrule 
      Body  & \texttt{Conv + LReLU + MP} & $7$ & 1 & $28/32$\\
      Encoder &\texttt{Conv + LReLU + MP} & $7$ & 1 & $32/48$\\
      & \texttt{Conv + LReLU + Global MP} & $7$ & 1 & $48/64$\\
      & \texttt{Conv} & $1$ & 1 & $64/16$\\
  \hline
  \midrule 
      View  & \texttt{Conv + LReLU + AP}  & $7$ & 1 & $28/32$\\
      Encoder & \texttt{Conv + LReLU + AP}   & $7$ & 1 & $32/48$\\
      & \texttt{Conv + LReLU + Global AP}  & $7$ & 1 & $48/64$\\
      & \texttt{Conv} & $1$ & 1 & $64/8$\\
  \hline
  \midrule 
      Decoder & \texttt{UpS + Conv + DO + LReLU}  & $7$ & 1 & $152/128$\\
      &\texttt{UpS + Conv + DO + LReLU} & $7$ & 1 & $128/64$\\
      &\texttt{UpS + Conv}  & $7$ & 1 & $64/30$\\
  \hline
  \bottomrule
\end{tabular}
\end{center}


\end{document}